\pdfoutput=1

\RequirePackage{fix-cm}

\documentclass[twocolumn]{svjour3}
\smartqed
\usepackage{cite}
\usepackage{graphicx}
\usepackage{epsfig}
\usepackage{epstopdf}
\usepackage{amsmath}
\usepackage{amssymb}
\usepackage{caption}
\usepackage{tabularx}
\usepackage[table]{xcolor}
\usepackage[normalem]{ulem}
\usepackage{algorithm}
\usepackage[noend]{algpseudocode}

\makeatletter
\def\BState{\State\hskip-\ALG@thistlm}
\makeatother

\definecolor{myGreen}{HTML}{33FF00}
\definecolor{myRed}{HTML}{FF3030}
\definecolor{myGrey}{HTML}{AA5555}
\definecolor{myWhite}{HTML}{FFFFFF}
\definecolor{maroon}{cmyk}{0,0.87,0.68,0.32}
\definecolor{petr}{HTML}{5555FF}
\definecolor{josef}{HTML}{FF3030}

\usepackage{blindtext}
\usepackage{booktabs}
\usepackage{url}
\usepackage {multirow}
\usepackage[font=footnotesize,skip=3pt,subrefformat=parens]{subcaption}

\usepackage[utf8]{inputenc}
\usepackage[T1]{fontenc}
\usepackage[pagebackref=true,breaklinks,colorlinks,citecolor=blue,linkcolor=blue,urlcolor=blue,hypertexnames=false]{hyperref}
\usepackage{url}
\usepackage{booktabs}
\usepackage{amsfonts}
\usepackage{nicefrac}
\usepackage{microtype}
\usepackage{xcolor}
\usepackage{multirow}
\usepackage{graphicx}

\usepackage{pifont}
\usepackage{makecell}
\usepackage{xspace}
\usepackage{amsmath}
\usepackage{colortbl}
\usepackage{wrapfig}

\def\ie{\textit{i.e.},\xspace}
\def\eg{\textit{e.g.},\xspace}

\newlength{\mylen}
\def\ourMethod{{HVG}\xspace}

\journalname{IJCV}

\begin{document}
\begin{sloppypar}
\title{Human Video Generation from a Single Image with 3D Pose and View Control}

\author{Tiantian Wang        \and
        Chun-Han Yao          \and
        Tao Hu   \and
        Mallikarjun Byrasandra Ramalinga Reddy \and
        Ming-Hsuan Yang \and 
        Varun Jampani
}

\institute{Tiantian Wang \at
            Electrical Engineering and Computer Science, University of California, Merced, CA 95343, United States.\\
            \href{twang61@ucmerced.edu}{twang61@ucmerced.edu}    
           \and
           Chun-Han Yao \at
           Stability AI, Los Angeles, CA 90067, United States.\\
            \href{chunhanyao@gmail.com}{chunhanyao@gmail.com} 
            \and
            Tao Hu \at
            Stability AI, Los Angeles, CA 90067, United States.\\
            \href{taohu.cs@gmail.com}{taohu.cs@gmail.com}
            \and
            Mallikarjun Byrasandra Ramalinga Reddy \at
            Stability AI, Los Angeles, CA 90067, United States.\\
            \href{mallik.jeevan@gmail.com}{mallik.jeevan@gmail.com}
            \and
            Ming-Hsuan Yang \at 
            Electrical Engineering and Computer Science, University of California, Merced, CA 95343, United States.\\
            \href{mhyang@ucmerced.edu}{mhyang@ucmerced.edu}
            \and
            Varun Jampani \at 
            Stability AI, Los Angeles, CA 90067, United States.\\
            \href{varunjampani@gmail.com}{varunjampani@gmail.com}
}   
\vspace{-5mm}
\date{Received: date / Accepted: date}

\maketitle

\begin{abstract}

\noindent Recent diffusion methods have made significant progress in generating videos from single images due to their powerful visual generation capabilities. However, challenges persist in image-to-video synthesis, particularly in human video generation, where inferring view-consistent, motion-dependent clothing wrinkles from a single image remains a formidable problem. In this paper, we present Human Video Generation in 4D (\textbf{\ourMethod}), a latent video diffusion model capable of generating high-quality, multi-view, spatiotemporally coherent human videos from a single image with 3D pose and view control. \ourMethod achieves this through three key designs:~i) \textbf{Articulated Pose Modulation} that captures the anatomical relationships of 3D joints via a novel dual-dimensional bone map and resolves self-occlusions across views by introducing 3D information;
ii) \textbf{View and Temporal Alignment}, ensuring multi-view consistency and alignment between a reference image and pose sequences for frame-to-frame stability; and
iii) \textbf{Progressive Spatio-Temporal Sampling} with temporal alignment to maintain smooth transitions in long multi-view animations. Extensive experiments on image-to-video tasks demonstrate that \ourMethod outperforms existing methods in generating high-quality 4D human videos from diverse human images and pose inputs.


\end{abstract}

\vspace{\mylen}
\section{Introduction}
\label{sec:intro}
\vspace{\mylen}

\begin{figure*}[t!]
    \centering    \includegraphics[width=0.9\linewidth]{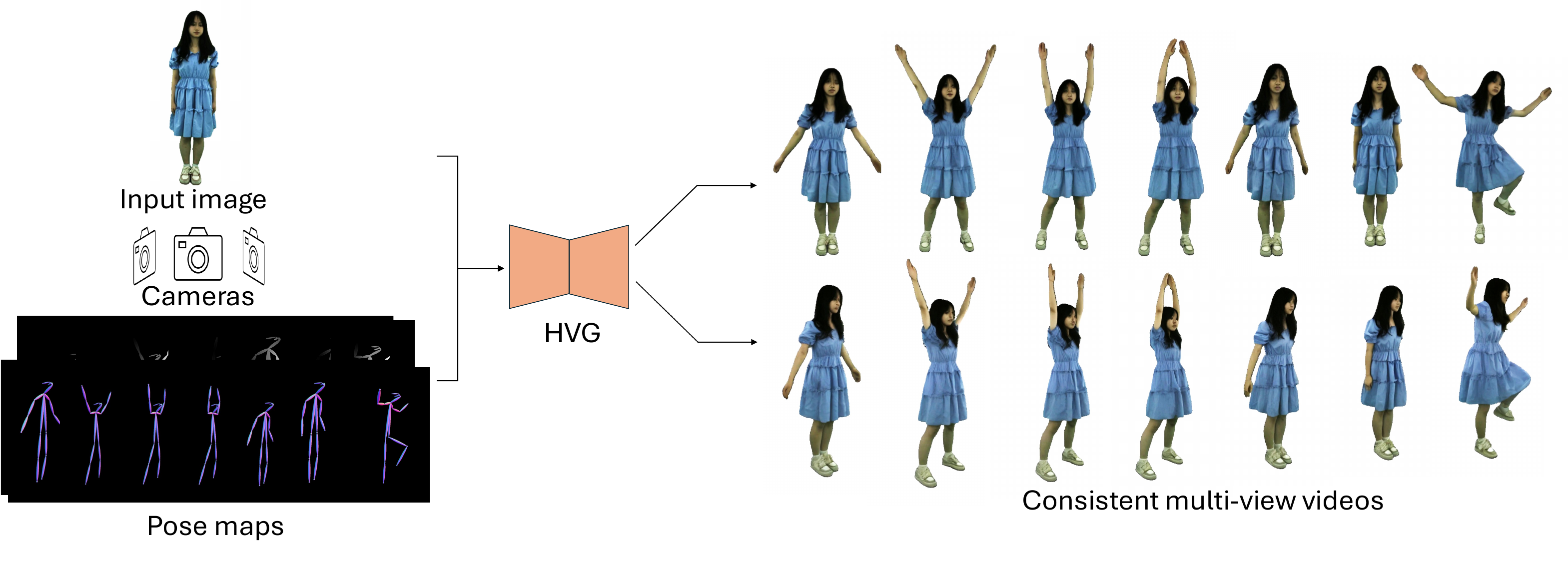}
    \caption{{\bf \ourMethod Overview.} \ourMethod is capable of generating consistent multi-view human videos from a single image, conditioned on the given multi-view pose sequences and camera poses.}
\label{fig:teaser}
\end{figure*}

Human video generation is widely used in animation, gaming, and virtual reality, enabling lifelike character synthesis for immersive storytelling and seamless virtual-real integration~\cite{lei2024comprehensive, shao2024human4dit, zhu2024champ, wang2023disco, icsik2023humanrf, zhang20254diffusion}. Recently, driven by the advances in generative image modeling with diffusion models~\cite{rombach2022high,ho2020denoising, Saharia2022PhotorealisticTD}, video diffusion models~\cite{voleti2022mcvd,lu2023vdt,kuang2024collaborative, singer2022make, guo2023animatediff, ho2022video, he2022latent, chen2023videocrafter1, chen2024videocrafter2, blattmann2023stable,voleti2024sv3d} have been proposed, which are capable of generating video sequences from single input images.

However, these video diffusion models~(\eg SVD~\cite{blattmann2023stable}) cannot effectively model articulated human motions, leading to suboptimal quality in dynamic human generation. 
To tackle this problem, some 2D human video animation methods propose pose-conditioned diffusion models, such as AnimateAnyone~\cite{hu2024animate} and MagicAnimate~\cite{xu2024magicanimate}, which deliver remarkable appearance consistency and motion stability. 
However, these methods work on only fixed camera viewpoints, restricting their ability to handle dynamic 3D scenarios that require multi-view human synthesis.

The limitations arise from their reliance on two error-prone driving signals: 2D skeletons~\cite{yang2023effective, cao2019openpose, guler2018densepose} and 3D clothless body meshes (\eg SMPL~\cite{loper2015smpl}, SMPL-X~\cite{pavlakos2019expressive}), leading to the following problems: \\
\textit{i)~2D skeleton based methods,}  such as Animate Anyone~\cite{hu2024animate} and Mimicmotion~\cite{zhang2024mimicmotion}, employ 2D-based method DWPose~\cite{yang2023effective} to guide pose transitions.
While effective in monocular settings, skeletons lack anatomical joint dependencies, such as hierarchical limb rotations and collision constraints.
This deficiency leads to implausible motions, including dislocated hips, hyperextended knees, or unnatural arm twists, when rendered from novel angles during dynamic actions such as turning.\\
\textit{ii)~SMPL-based methods,} such as Champ~\cite{zhu2024champ}, parameterize human bodies using  3D clothless meshes. 
However, SMPL’s topology oversimplifies character-specific geometry, failing to represent loose clothing, accessories, or unique body shapes.
This results in shape leakage, where distortions, warped garment edges, or inconsistent limb proportions emerge under multi-view synthesis~\cite{pavlakos2019expressive}.

To advance multi-view synthesis, Human4DiT~\cite{shao2024human4dit} introduces a 4D Diffusion Transformer that generates multi-view human videos using SMPL-derived normal maps as driving signals. This method leverages a view attention mechanism to align spatial features across multiple perspectives. However, SMPL-derived normal maps introduces shape inaccuracies due to the simplified geometric representation of human bodies. Additionally, the view attention mechanism, while enhancing cross-view consistency, incurs significant computational overhead, limiting its scalability and efficiency in 4D video generation. These challenges highlight the need for more expressive human representations and optimized attention mechanisms to improve both shape fidelity and computational efficiency.

To address these limitations, we introduce Stable Human Video in 4D~(\ourMethod), 
a latent diffusion model that synthesizes dynamic 4D human videos with multi-view and multi-frame consistency, which is demonstrated in Figure~\ref{fig:teaser}. 
\ourMethod incorporates three key innovations:

\noindent \textbf{Articulated Pose Modulation.} 
We propose a novel dual-dimensional bone map as driving signal representations, constructed by connecting 3D skeletal joints~\cite{pavlakos2019expressive} with ellipsoids to model anatomical relationships. This 3D structure is then projected onto 2D planes, which ensures structural fidelity, resolves occlusions (\eg limbs crossing in novel views), and mitigates shape distortions.  By encoding joint thickness, orientation, and collision constraints, ellipsoids retain volumetric cues absent in 2D skeletons while avoiding SMPL’s geometric oversimplification.  
When projected, they preserve depth-ordering and proportional relationships, preventing shape leakage (\eg distorted clothing) and maintaining anatomically consistent, view-invariant geometry. The 2D bone maps are independently processed by two pose modulators, each capturing complementary aspects of spatial and temporal information, which are then fused through a cross-attention mechanism.

\noindent \textbf{View and Temporal Alignment.}
\ourMethod integrates a lightweight view alignment strategy, applying human-centric centering to align multi-view attention for efficient cross-view correspondence computation, rather than relying on computationally intensive viewpoint-spatial 3D attention calculations.  
Additionally,~\ourMethod incorporates temporal alignment with the reference image and guided pose sequences to enhance frame-to-frame stability.

\noindent \textbf{Multi-View Video Generation.}
We introduce a progressive spatial-temporal sampling technique with temporal alignment, operating with alternating maximizing temporal windows~(\eg 24 frames) and view windows~(\eg 6 views).  
This enhances viewpoint coverage, preserves temporal consistency, and optimizes computational efficiency.

\ourMethod is evaluated on single image human video generation tasks under novel views and novel poses, where \ourMethod achieves superior results in motion accuracy, shape preservation, and temporal coherence compared to existing methods. 
By bridging pose-guided animation, multi-view synthesis, and long-sequence generation, \ourMethod advances the state of multi-view human video synthesis.

\vspace{\mylen}
\section{Related Work}
\label{sec:related}
\vspace{\mylen}

\noindent {\bf Diffusion-Based Image and Video Synthesis.}
Diffusion-based approaches have excelled in generating high-quality images~\cite{ho2020denoising, song2020score, rombach2022high, zhang2023adding, wang2024instantid} and videos~\cite{lei2024comprehensive, shao2024human4dit, zhu2024champ, wang2023disco, icsik2023humanrf, zhang20254diffusion}.
Latent diffusion models, such as Stable Diffusion v3.5~\cite{aidrawing}, efficiently generate high-resolution images by operating in a compact latent space instead of raw pixels.

Video synthesis demands modeling both spatial structures and temporal coherence.  
Diffusion-based advances achieve this by integrating temporal layers into pre-trained image models~\cite{singer2022make, guo2023animatediff, wu2023tune, wang2024magicvideo, blattmann2023stable} or utilizing transformer architectures~\cite{yan2021videogpt, yu2023magvit} for video generation.
For example, Stable Video Diffusion~(SVD)~\cite{blattmann2023stable}, built on Latent Diffusion Models~(LDM)~\cite{rombach2022high}, extends text-to-image diffusion by incorporating temporal layers and fine-tuning on large-scale, high-quality video datasets, enabling an open-source framework for high-fidelity video synthesis.  
In this work, we advance SVD for pose-guided 4D human video generation, leveraging its pre-trained generative capability to improve motion control and realism.

\noindent {\bf Pose-Guided Human Video Generation.}
Human video generation transforms static images into realistic, temporally coherent videos.  
Recent advances leverage human pose as a guiding signal, categorized into keypoint-based and SMPL-based methods~\cite{wang2023disco, xu2024magicanimate, hu2024animate, hu2025animate, zhu2024champ, shao2024human4dit, wang2023disco}.

Keypoint-guided methods use sparse skeletal representations, such as OpenPose~\cite{cao2019openpose} and DWPose~\cite{yang2023effective}, to control motion.
Animate Anyone~\cite{hu2024animate} enhances these approaches by employing a UNet-based ReferenceNet to extract features from reference images while integrating pose information through a lightweight convolutional pose guider.
MagicAnimate~\cite{xu2024magicanimate} follows a similar framework but incorporates a ControlNet tailored for DensePose~\cite{guler2018densepose}, offering more precise pose guidance than OpenPose~\cite{cao2019openpose}.  
However, 2D skeletons lack explicit modeling of inter-limb spatial relationships and hierarchical dependencies, leading to implausible motions, particularly in dynamic 3D scenarios.

SMPL-guided methods integrate 3D parametric meshes for richer geometry~\cite{wang2023disco, zhu2024champ, shao2024human4dit}.  
Champ~\cite{zhu2024champ} builds on Animate Anyone’s framework but adopts SMPL guidance to model surface deformations, spatial relationships (\eg occlusions), and contours.  
Although 3D mesh representations like SMPL effectively model human bodies, they compromise generalizability across characters and may introduce shape leakage due to their dense representation.

In contrast, \ourMethod introduces bone maps—connecting 3D SMPL joints with ellipsoids, projected into 2D—to maintain structural fidelity, resolve occlusions, and reduce shape distortions, while aligning views and frames for multi-view, multi-frame video generation.

\noindent {\bf Multi-View Video Generation.}
Long video generation aims to extend video duration while preserving temporal coherence. Existing methods, such as AnimateAnyone~\cite{hu2024animate} and MimicMotion~\cite{zhang2024mimicmotion}, employ progressive latent fusion—a training-free technique that integrates seamlessly into the denoising process of latent diffusion models during inference. Motivated by this approach, we introduce a novel spatio-temporal sampling strategy that applies progressive latent fusion across both view and temporal dimensions.  Specifically, we segment multi-view pose sequences into overlapping parts during inference to enhance consistency. 
In contrast, Human4DiT employs a multi-view multi-frame sampling strategy by partitioning views and frames into non-overlapping segments. Our proposed method demonstrates improved multi-view, multi-frame consistency.

\vspace{\mylen}
\section{Methodology}\label{sec:method}
\vspace{\mylen}

\begin{figure*}[tp]
    \centering    \includegraphics[width=\linewidth]{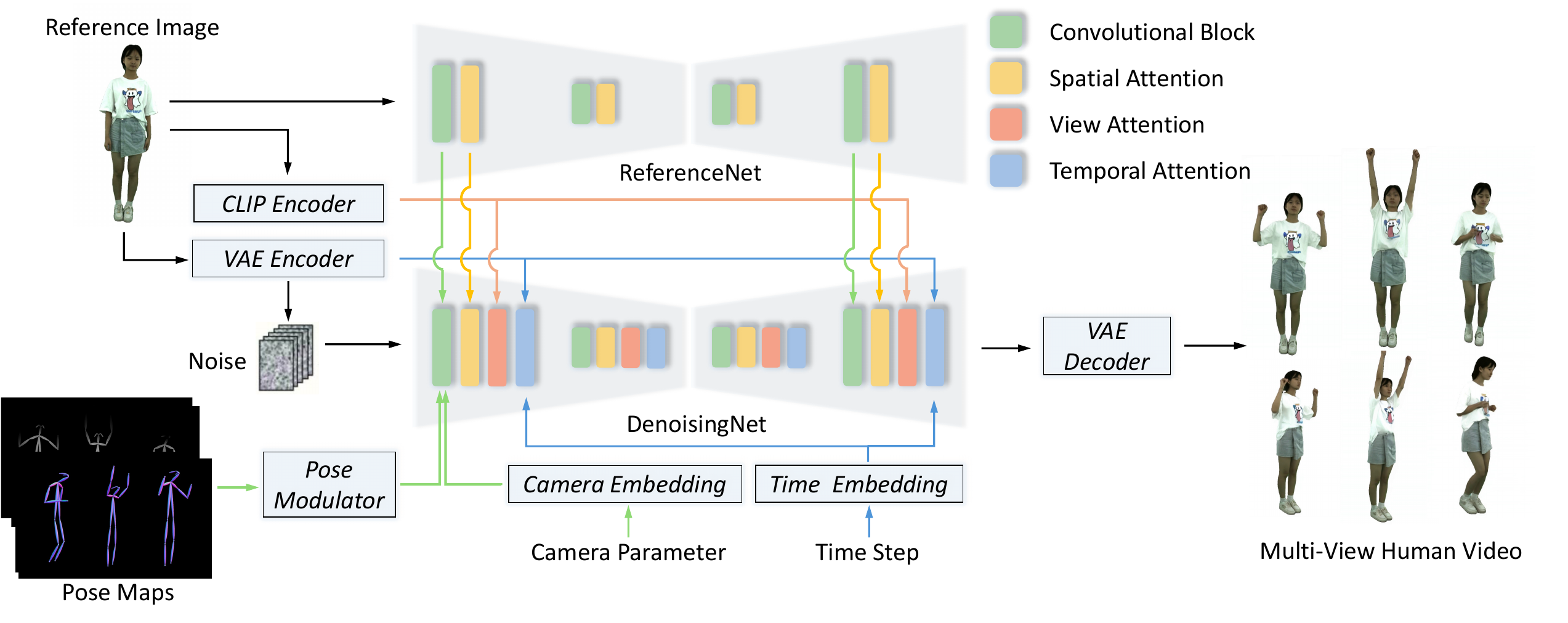}
    \vspace{-5mm}
\caption{\textbf{Framework of \ourMethod.} The bone map sequence is processed by the pose modulator, followed by the denoising process of DenoisingNet, which integrates camera parameters via camera embedding and time steps to generate a multi-view human video. DenoisingNet consists of convolutional blocks, spatial attention, view attention, and temporal attention to capture temporal and spatial correspondences. The reference image contributes in three key ways: First, ReferenceNet extracts fine-grained details to enhance spatial attention. Second, semantic features are captured through the CLIP Encoder for convolutional blocks and view attention and are fused with multi-frame noise. Third, the VAE Encoder processes reference image features for temporal attention. }
\label{fig_overall}
\end{figure*}

Given a reference image of a person~$\mathcal{I}_{ref}$, a sequence of driving motion frames~$\left\{\mathcal{B}_i\right\}_{i=1}^{T}$, and camera parameters~$\left\{\mathcal{C}_i\right\}_{i=1}^{V}$, where $T$ and $V$ denote the total number of video frames and viewpoints, respectively, our goal is to generate a sequence of photo-realistic video frames~$\left\{\mathcal{I}_t^v \mid t=1, \dots, T, \; v=1, \dots, V \right\}$ depicting the person performing the specified motion from given viewpoint, maintaining temporal and multi-view consistency.

In this section, we first introduce the multi-dimensional bone maps, followed by a detailed description of the network structure.  
Finally, we present a progressive spatio-temporal sampling strategy for multi-view long video generation.

\vspace{\mylen}
\subsection{Dual-Dimensional Bone Map}
\label{subsec_bone_map}
\vspace{\mylen}

Given the SMPL parameters of a human~\cite{loper2015smpl}, we calculate a dual-dimensional bone map \( \mathcal{B} \) as pose guidance, consisting of a depth map \( \mathcal{D} \) and normal map \( \mathcal{N} \).

\noindent \textbf{Joint Extraction.}  
Given a posed SMPL-X mesh~\cite{pavlakos2019expressive}, we first extract \( N = 23 \) joint coordinates~\( \mathbf{J} = \{\mathbf{j}_n \in \mathbb{R}^3\}_{n=1}^{N} \), where each joint pair~\( (m, n) \) corresponds to a skeletal bone segment.

\noindent \textbf{Ellipsoid Modeling.}
For each bone segment between joints $\mathbf{j}_m$ and $\mathbf{j}_n$, we model its volumetric structure with a 3D ellipsoid $\mathcal{E}_{mn}$ centered at $\mathbf{c}_{mn} = \frac{\mathbf{j}_m + \mathbf{j}_n}{2}$. 
The ellipsoid's orientation $\mathbf{R}_{mn} \in \mathbb{R}^{3\times 3}$, a 3D rotation matrix, aligns its major axis with the bone direction $\mathbf{d}_{mn} = \mathbf{j}_n - \mathbf{j}_m$:
\begin{equation}
    \mathbf{R}_{mn} = \text{Align}\left(\frac{\mathbf{d}_{mn}}{\|\mathbf{d}_{mn}\|_2}, \mathbf{e}_z\right)
\end{equation}
where $\text{Align}(\mathbf{v}, \mathbf{u})$ computes the rotation matrix aligning vector $\mathbf{v}$ with $\mathbf{u}$, and $\mathbf{e}_z = (0,0,1)$ is the canonical upward axis. 
The ellipsoid radii $\mathbf{r}_{mn} = (r_x, r_y, r_z)$ are set proportionally to anthropometric measurements, with $r_z = \frac{1}{2}\|\mathbf{d}_{mn}\|$ along the bone axis, and $r_x = r_y$ determined by joint-specific thickness, \eg 2.5 cm for wrists~\cite{805368, bogo2017dynamic}.

\noindent \textbf{Pose Map Rendering.}
We render the 3D ellipsoidal structure onto 2D images via perspective projection~\cite{hartley2003multiple} implemented via PyTorch3D~\cite{ravi2020pytorch3d}, producing two complementary pose motion maps:

\textit{Depth Map}: 
Encodes z-ordering for occlusion resolution by representing the distance of each pixel to the origin in camera space.
This provides 3D spatial information, capturing the relative positioning of body parts in the 2D image.

\textit{Normal Map}: Preserves body surface orientation by capturing the direction of each point on the body. 
To maintain multi-view consistency, we transform the orientation from world space to camera space.

\noindent
\textbf{Advantages.}
Bone map addresses two limitations.
\textit{i)~Skeleton fragility:} ellipsoid projections encode limb volume/spatial occupancy to prevent implausible intersections; 
\textit{ii)~SMPL rigidity:} parametric modeling preserves clothing/accessories by decoupling shape from pose. 
Perspective projection inherently maintains anatomical fidelity across views, eliminating mesh-induced shape leakage.

\begin{figure*}[t!]
    \centering    \includegraphics[width=0.8\linewidth]{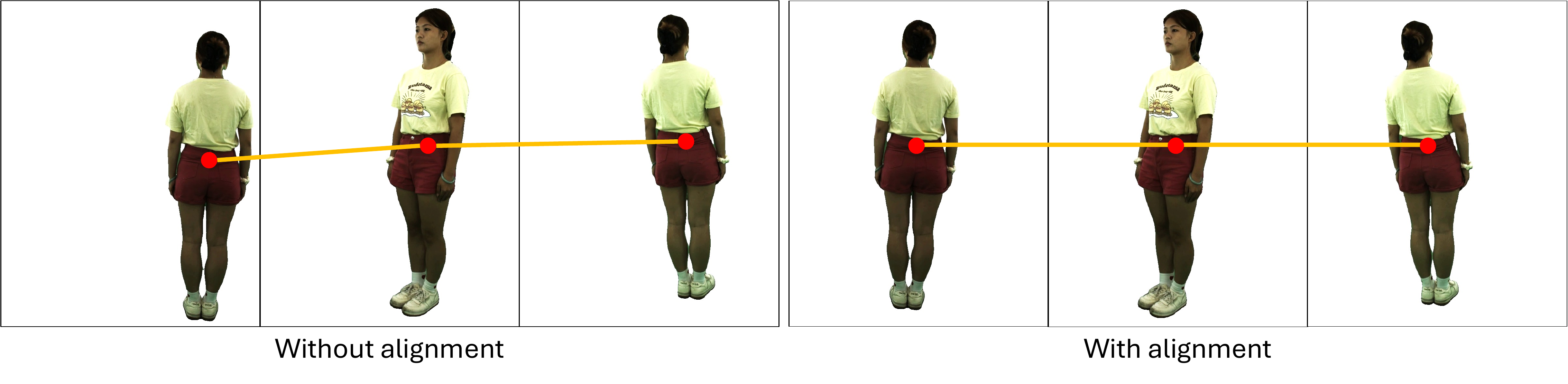}
    \caption{{\bf Human position alignment.} The left and right figures display the human subject across different views, both without and with alignment. After alignment, the human subject is positioned consistently in the same location across views.}
\label{fig:human_alignment}
\end{figure*}

\vspace{\mylen}
\subsection{Network Architecture}\label{sec_network}
\vspace{\mylen}

Figure~\ref{fig_overall} presents the \ourMethod.  
The main network is the denoising UNet~(DenoisingNet), designed similarly as SV4D~\cite{xie2024sv4d} and initialized using the pre-trained SVD-xt~\cite{blattmann2023stable} model.  
It takes noise as input and generates multi-view human videos, conditioned on a reference image, pose map sequence, and camera parameters.  
Each layer of DenoisingNet is equipped with 3D convolution, spatial attention, view attention, and temporal attention blocks to effectively capture spatio-temporal correspondences.  
The reference image is processed through three modules to preserve human identity: ReferenceNet, a VAE encoder, and a CLIP encoder, whose outputs are incorporated into the DenoisingNet.

\ourMethod incorporates two key components: 
i)~a pose modulator that encodes our multi-dimensional bone maps as driving signals;  
ii)~an efficient view attention layer.

\noindent \textbf{Pose Modulator.}
The pose modulator extracts pose information from our dual-dimensional bone map, consisting of depth and normal maps. 
It processes input dual pose maps through the Normal Map Guider and the Depth Map Guider, each utilizing four convolutional layers with \( 4\times4 \) kernels, \( 2\times2 \) strides, and channel dimensions of \( 16, 32, 64, 128 \), similar to the condition encoder in ControlNet~\cite{zhang2023adding}.  
These processed pose features are then fed into a cross-attention layer, which incorporates spatial and structural information from our bone map.
The output of the cross-attention layer is integrated into the first convolutional block of the DenoisingNet, which serves as the animated guidance.

\noindent 
\textbf{Efficient View Attention Layer.} 
Each sampling layer consists of four components: \ie a convolutional block, spatial attention, view attention, and temporal attention.  
We elaborate only on our efficient view attention as the other layers function the same as in SVD~\cite{blattmann2023stable}.

View attention is integrated to learn cross-view consistency in multi-view settings.  
In multi-view images, human positions vary significantly due to perspective distortion and changes in camera viewpoints.  
This poses a challenge for cross-view attention mechanisms, as the relative pixel locations of corresponding body parts shift considerably across views.  
For example, the head may appear at the center of the image in one view but shift to the upper left in the next.

Human4DiT~\cite{shao2024human4dit} addresses this by introducing a 3D view attention mechanism that jointly processes viewpoint and 2D spatial dimensions.  
This enables the model to learn global correlations across views while accounting for spatial variations.  
However, this approach significantly increases computational complexity, as its 3D view attention layer processes tokens~\( \mathbf{z}_t^v \in \mathbb{R}^{T \times (V \times H \times W) \times C} \) across views~(\( V \)) and spatial dimensions~(\( H \times W \)).  
As a result, the feasible window size for viewpoints is constrained, limiting the ability to effectively capture cross-view correlations.
 
To overcome this problem, we propose a simple and efficient view alignment strategy that eliminates costly 3D attention.
Instead of directly learning cross-view correlations, we align the human subject to a consistent position across views using the human body center to eliminate spatial variation.
Subsequently, only 2D attention is adopted to learn cross-view correlations.  
This strategy not only reduces the computational load of the view attention layer but also maintains spatial consistency across views, optimizing both accuracy and efficiency in multi-view processing.

Specifically, we first extract the pelvis joint from the SMPL model and project it into the 2D image space.  
We then shift all human figures to a consistent position (images are cropped). Figure~\ref{fig:human_alignment} provides visual examples of human alignment before and after our proposed adjustment.

\noindent 
\textbf{Reference Module.} 
The reference image is integrated into the model through three pathways to ensure consistent human identity representation. First, we use CLIP \cite{radford2021learning} to extract an image embedding, which is fed into each U-Net block via a cross-attention mechanism. This approach enables the network to maintain spatial and temporal consistency across new views and frames while preserving the semantic context of the input video. Second, a VAE encoder encodes the image features, which are concatenated with the noise latents before being passed into the U-Net. This allows the network to capture finer identity characteristics from the reference image. Third, we employ the ReferenceNet architecture from~\cite{hu2024animate} to extract appearance features from the reference image. To reduce computational complexity, our framework integrates these features exclusively into the downblock and midblock of the DenoisingNet decoder using spatial attention.

\noindent
\textbf{Embeddings.} 
We use camera pose and frame index embeddings to distinguish between different views and frames. For the camera pose, we use the rotation matrix, setting the rotation of the first view as an identity matrix. Subsequent views’ rotations are calculated relative to this initial view, allowing for a consistent reference across frames. Both the camera pose and frame index embeddings utilize sinusoidal positional encoding to capture the inherent spatial and temporal relationships.
The camera pose embeddings are concatenated, then linearly transformed, and integrated into the noise timestep embedding. %
Frame index embeddings, on the other hand, are incorporated into the temporal layer and fed into the model before applying temporal attention.

\begin{figure}[t]
    \centering
    \includegraphics[width=0.48\textwidth]{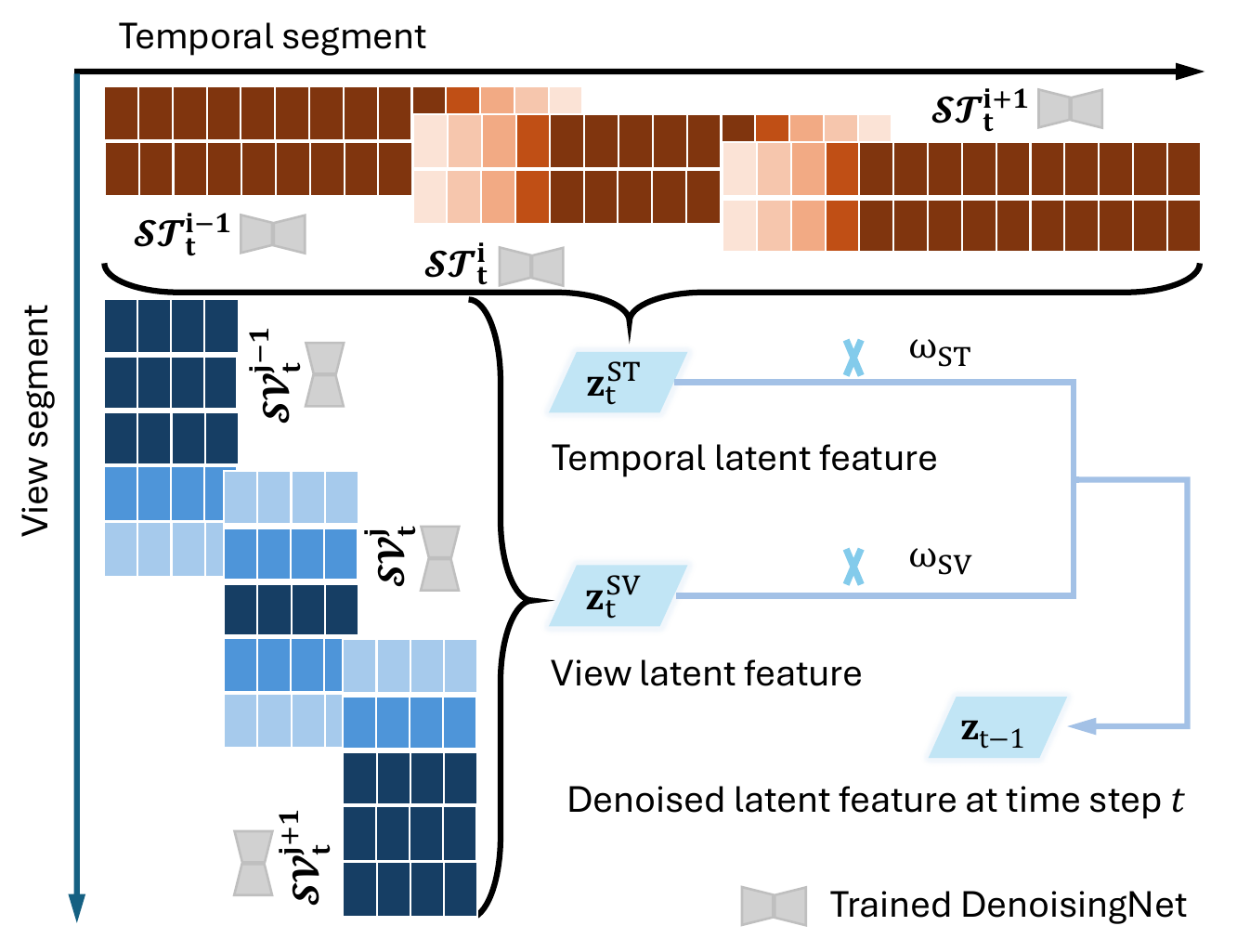}
    \caption{\textbf{Illustration of spatio-temporal sampling.} To generate a multi-view long video, we divide the sequence into overlapping segments along both the temporal and view dimensions, denoted as~\{\(\mathcal{ST}_t^i\)$|i=1,2,\ldots\}$ and~\{\(\mathcal{SV}_t^j\)$|j=1,2,\ldots\}$. These segments are independently aggregated to form long-range latent representations ~\(\mathbf{z}_{t}^{\text{ST}}\) and~\(\mathbf{z}_{t}^{\text{SV}}\) in the temporal and view dimensions at each timestep $t$. At each timestep \( t \), the denoised temporal latent~\( \mathbf{z}_{t}^{\text{ST}} \) and view latent~\( \mathbf{z}_{t}^{\text{SV}} \) are combined through a learned weighting strategy to produce the updated latent feature \( \mathbf{z}_{t-1} \). 
    Repeating this denoising process until \( t=1 \) yields \( \mathbf{z}_0 \), which is then decoded to synthesize the final long multi-view human video.}
    \label{fig:tv_sampling} 
\end{figure}

\vspace{\mylen}
\vspace{\mylen}
\subsection{Progressive Spatio-Temporal Sampling}
\label{sec:section_3}
\vspace{\mylen}

Generating videos with camera view and temporal variations is computationally intensive~\cite{shao2024human4dit}.
To address this, we introduce a progressive spatio-temporal sampling strategy to synthesize human videos while maintaining both temporal and view consistency.

To generate a multi-view long video, we sample in both temporal and view dimensions, as shown in Figure~\ref{fig:tv_sampling}. 
In the temporal dimension, the video is partitioned into multiple overlapping long segments, each with \( T_{\text{long}} \) frames and \( N_{\text{short}} \) views,  using a sliding window with \( T_{\text{ol}} \) overlapped frames.
In the view dimension, the video is partitioned into segments, each containing \( T_{\text{short}} \) frames and \( N_{\text{long}} \) viewpoints, with  \( N_{\text{ol}} \) overlapped views. 
At each denoising timestep \( t \), we process each temporal and view segment as follows:

\textbf{i)~Temporal Segment Processing}. As depicted in Figure~\ref{fig:tv_sampling}, the \( i \)-th temporal segment~\( \left\{\mathcal{ST}_t^i\right\} \) is denoised using the trained DenoisingNet, conditioned on the reference image, bone map subsequence, and camera viewpoint.  
To ensure smooth transitions, we apply weighted fusion to the latent features of each frame in~\( \left\{\mathcal{ST}_t^i\right\} \).  
Non-overlapping frames remain unchanged, while overlapping frames from adjacent segments~\(\left( \left\{\mathcal{ST}_t^{i-1}\right\}, \left\{\mathcal{ST}_t^{i+1}\right\} \right)\) are assigned higher weights if they are closer to the \(i\)-th segment. 
After merging all the segments, we obtain the temporal-dimension long video latent feature~\( \mathbf{z}_{t}^{\text{ST}} \) at timestep $t$.

\noindent \textbf{Temporal Alignment.}
We introduce temporal alignment to enhance frame-to-frame stability by ensuring the consistency between the reference image and pose guidance. This is achieved by aligning the human position in the reference image with the center of the first frame from the pose sequences to a consistent position.
This minimizes flickering artifacts and pose discontinuities, resulting in smoother and more coherent videos.

\textbf{ii)~View Segment Processing.}  
As shown in Figure  \ref{fig:tv_sampling}, each view segment~\( \left\{\mathcal{SV}_t^j\right\} \) is denoised using the trained DenoisingNet, conditioned on the reference image, bone maps, and camera viewpoints.  
After merging all denoised view segments, we obtain the view-dimension video latent feature~\( \mathbf{z}_{t}^{\text{SV}} \) at timestep $t$.

At each timestep \( t \), the denoised temporal-dimension latent feature~\( \mathbf{z}_{t}^{\text{ST}} \) and view-dimension latent feature~\( \mathbf{z}_{t}^{\text{SV}} \) are weighted combined, 
which produce the final denoised latent feature \( \mathbf{z}_{t-1} \).  
We compute the \( \mathbf{z}_0 \) by applying the above denoising process until timestep \( t=1 \),  which is then decoded to generate the final long multi-view human video.

\vspace{\mylen}
\section{Experiments}
\label{sec:experiment}
\vspace{\mylen}

\begin{table*}[t]
\centering
\footnotesize
\renewcommand\arraystretch{.1}
\resizebox{0.9\linewidth}{!}{%
    \begin{tabular}{lccccccc}
        \toprule
        \textbf{Method} & \textbf{FID $\downarrow$} & \textbf{SSIM $\uparrow$} & \textbf{PSNR $\uparrow$} & \textbf{LPIPS $\downarrow$} & \textbf{L1 $\downarrow$} & \textbf{FID-VID $\downarrow$} & \textbf{FVD $\downarrow$} \\
        \midrule
        MagicAnimate~\cite{xu2024magicanimate}  &363.1 &0.756 &8.227 &0.455 &2.8e-04&161.3 &1323.\\
        AnimateAnyone~\cite{hu2024animate} &117.5 &0.868 &17.48 &0.179 &4.6e-05&81.15 &660.7 \\
        Champ~\cite{zhu2024champ} &126.1 &0.886 &17.74 &0.108 &2.8e-05&34.00 &410.2  \\
        MimicMotion~\cite{zhang2024mimicmotion}&223.6 &0.905 &19.59 &0.082 &3.2e-05&24.55 &271.1  \\
        AniGS~\cite{qiu2024anigs}&101.5 &0.900 &18.97 &0.084 &3.0e-05&28.62 &262.0  \\
        LHM~\cite{qiu2025lhm}&81.60 &0.907 &19.82 &0.075 &1.8e-05&25.80 &248.4  \\
        \arrayrulecolor{black!30}\midrule
        AnimateAnyone*~\cite{hu2024animate}  &106.5   &0.892 &18.39  &0.093 &2.6e-05&30.74 &295.2 \\
        Champ*~\cite{zhu2024champ}  &88.31   &0.904 &19.35  &0.079 &2.2e-05&27.42 &257.4 \\
        \rowcolor{gray!20} \ourMethod &59.35 &0.923 &22.13 &0.057 &1.5e-05&13.97 &152.1 \\
        \arrayrulecolor{black}\bottomrule
    \end{tabular}
}
\caption{{\bf Comparison for novel view synthesis on 3D scans with image \& video metrics.} $*$ indicates that we fine-tuned the model on our dataset. Our method outperforms prior work and demonstrates clear effectiveness in learning accurate 3D viewpoint transformations.}
\label{tab:3d_scan}
\end{table*}

\begin{figure*}[tp]
    \centering
\includegraphics[width=0.95\linewidth]{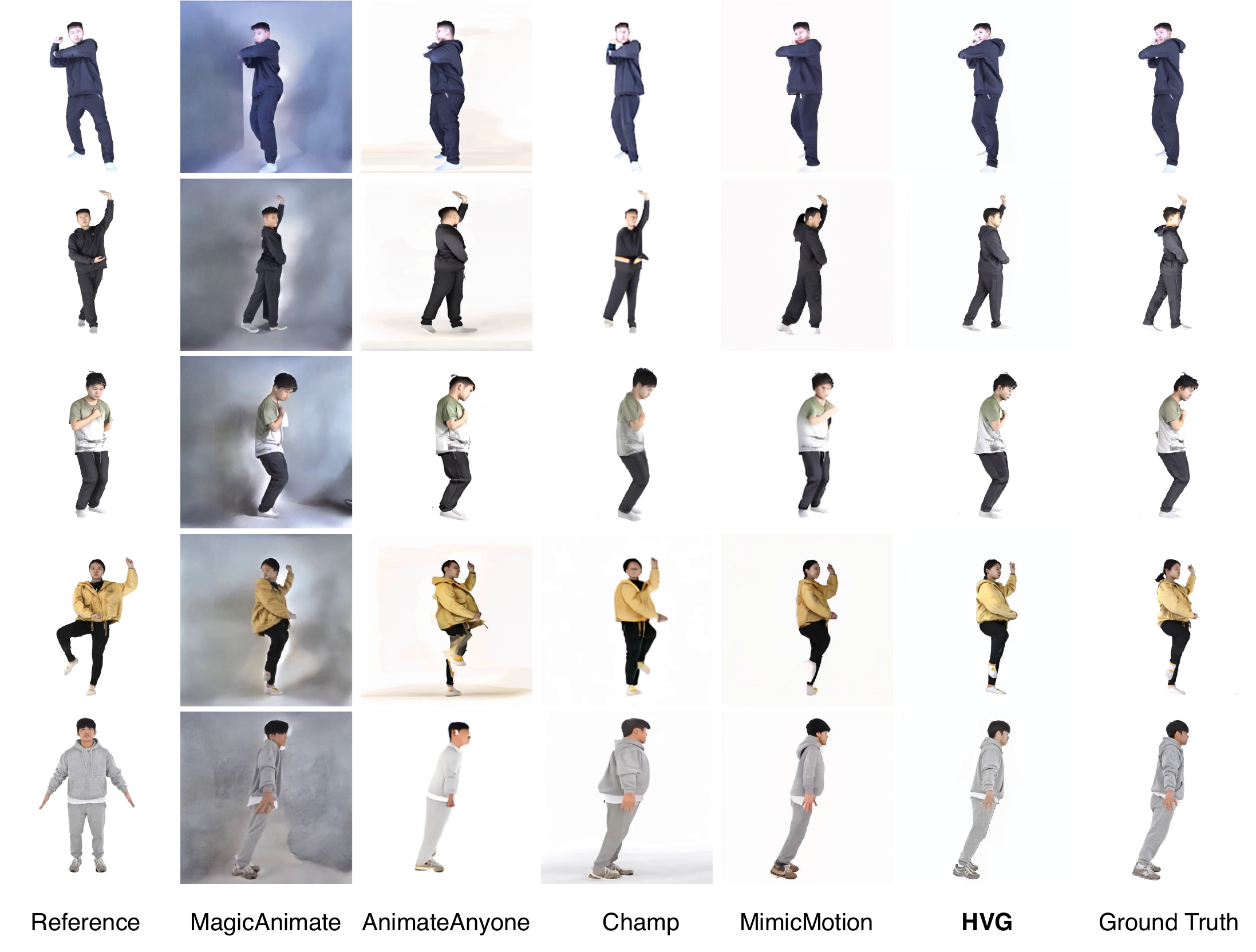}
        \caption{{\bf Novel-view results from single-view images.} \ourMethod (ours) generates novel-view images with higher anatomic fidelity, faithfulness to input images, and consistency of geometry and texture.}
\label{fig:fig_visual_multi-view}
\end{figure*}

\begin{table*}[tp]
\centering
\footnotesize
\renewcommand\arraystretch{.1}
\resizebox{0.9\linewidth}{!}{%
\begin{tabular}{lccccccc}
\toprule
\textbf{Method} & \textbf{FID $\downarrow$} & \textbf{SSIM $\uparrow$} & \textbf{PSNR $\uparrow$} & \textbf{LPIPS $\downarrow$} & \textbf{L1 $\downarrow$} & \textbf{FID-VID $\downarrow$} & \textbf{FVD $\downarrow$} \\
\midrule
MagicAnimate~\cite{xu2024magicanimate} &219.1 &0.722 &8.173 &0.478 &1.8e-04&211.0 &1197  \\
AnimateAnyone~\cite{hu2024animate}  &154.1 &0.843 &16.71 &0.218 &3.8e-05&96.12 &718.4  \\
Champ~\cite{zhu2024champ}  &158.2 &0.858 &16.29 &0.145 &2.7e-05&53.88 &468.0  \\
MimicMotion~\cite{zhang2024mimicmotion} &120.5 &0.876 &18.26 &0.119 &2.8e-05&34.98 &291.0 \\
AniGS~\cite{qiu2024anigs}  &105.3 &0.871 &17.36 &0.127 &2.9e-05&36.31 &299.3\\    
LHM~\cite{qiu2025lhm}  &103.7 &0.884 &18.43 &0.110 &1.8e-05&28.86 &258.8\\
\arrayrulecolor{black!30}\midrule
AnimateAnyone*~\cite{hu2024animate}  &124.9  &0.861 &17.14  &0.141 &2.5e-05&49.28 &327.4\\
Champ*~\cite{zhu2024champ}  &113.8  &0.876 &17.82  &0.119 &2.1e-05 &32.47 &275.6\\
\rowcolor{gray!20} \ourMethod &84.74 &0.901 &20.90 &0.092 &1.3e-05&20.05 &177.4 \\
\arrayrulecolor{black}\bottomrule
\end{tabular}
}
\caption{{\bf Comparison for novel view and novel pose synthesis on multi-view videos.} $*$ indicates that we fine-tuned the model on our dataset. Our approach HVG surpasses existing methods by effectively addressing both camera viewpoint differences and human motion variations at the same time.}
\label{tab:3d_scan_quantitative}
\end{table*}

\begin{figure*}[h]
    \centering
\includegraphics[width=0.95\linewidth]{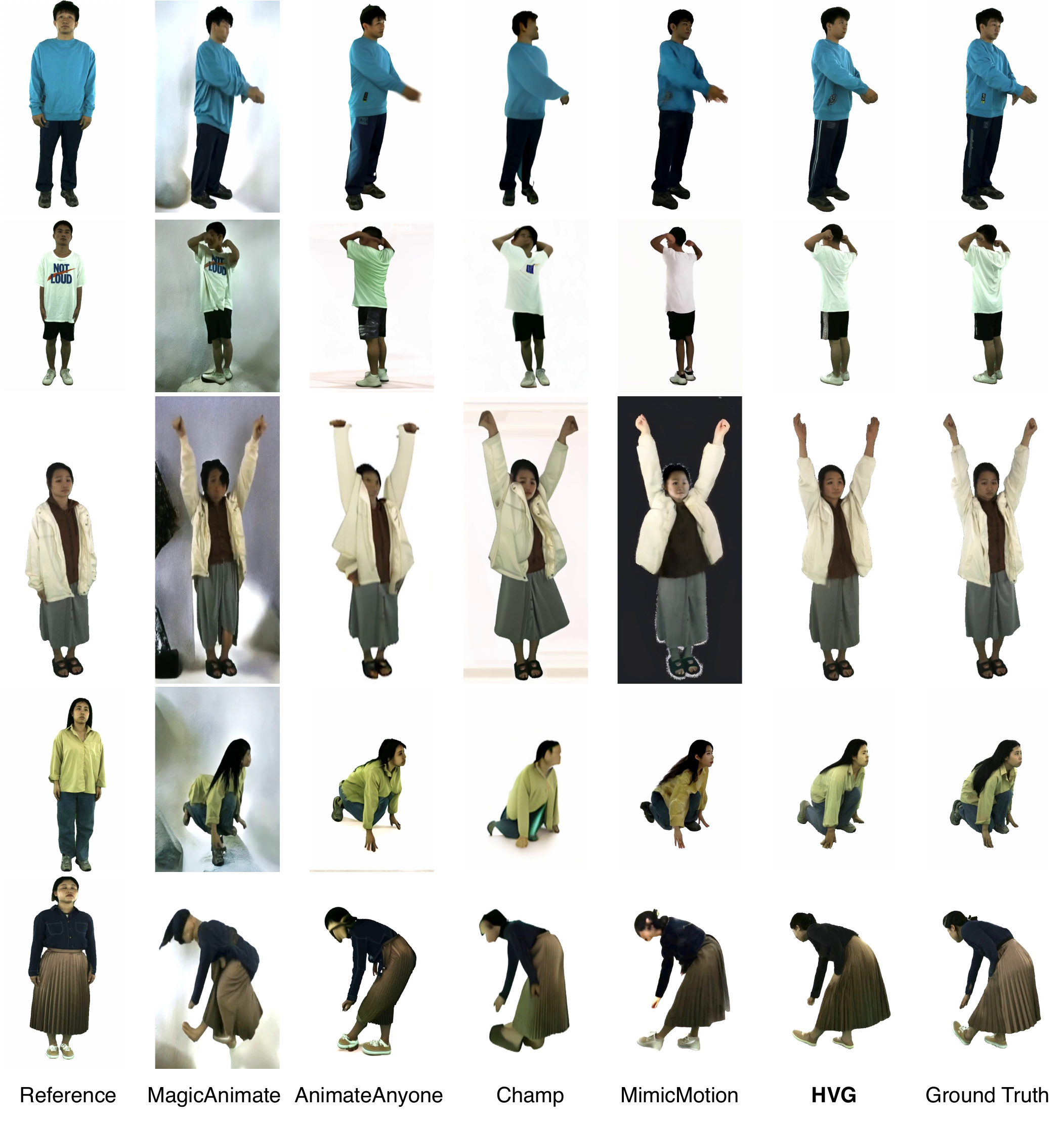}
    \vspace{-5pt}
        \caption{{\bf Novel-view novel-pose results from single-view images.} \ourMethod generates novel-view images with higher anatomic fidelity, faithfulness to inputs, and consistency of geometry and texture.}
    \vspace{-2pt}
\label{fig:fig_visual}
\end{figure*}

\noindent 
\textbf{Implementation.} 
Training consists of two phases: first, we train a multi-view model for 60,000 steps with a batch size of 2, where each batch contains 48 views. 
Next, we initialize multi-view multi-frame model with the first-stage model and further finetune it for 50,000 steps with a batch size of 1. 
Each batch includes 72 images, accommodating varying numbers of frames and views.
During inference, we employ a progressive spatio-temporal sampling method to generate long multi-view videos. 
Our experiments are conducted using 32 NVIDIA H100 GPUs. 
We process individual multi-view images or frames by sampling, resizing, and cropping them to a uniform resolution of 576$\times$576 pixels.

\noindent 
\textbf{Dataset.} 
Our model is trained using a combination of several datasets, including THuman2.0~\cite{yu2021function4d}, THuman2.1~\cite{yu2021function4d}, CustomHuman~\cite{ho2023learning}, 2K2K~\cite{han2023high}, and MVHumanNet~\cite{xiong2024mvhumannet}. 
The first four datasets contain approximately 5,000 3D scans, which are used to render 360-degree novel views for training the multi-view model.
The MVHumanNet dataset is employed to train our multi-view multi-frame model with 5000 multi-view videos.

The training set for the multi-view model includes the first 500 videos from THuman2.0, the first 1,900 videos from THuman2.1, the first 625 videos from CustomHuman, and the first 1,980 videos from 2K2K. To avoid subject overlap between training and inference, we remove any videos featuring the same subjects as those in the training set and randomly select 50 videos from the remaining pool for inference.
For the multi-view multi-frame setting, the test set consists of 25 randomly selected 360-degree multi-view videos.

\noindent 
\textbf{Metrics.}  
We evaluate model performance based on both single-frame quality and video fidelity~\cite{wang2023disco}.
For single-frame evaluation, we adopt FID~\cite{heusel2017gans}, SSIM~\cite{wang2004image}, PSNR~\cite{hore2010image}, LPIPS~\cite{zhang2018unreasonable}, and L1 error.
For video fidelity, we use FFID-FVD~\cite{balaji2019conditional} and FVD~\cite{unterthiner2018towards} to measure temporal and multi-view coherence.

\noindent
\textbf{Baselines.}  
We compare \ourMethod with state-of-the-art pose-guided human video generation methods, including MagicAnimate~\cite{xu2024magicanimate}, AnimateAnyone~\cite{hu2024animate}, Champ~\cite{zhu2024champ}, and MimicMotion~\cite{zhang2024mimicmotion}.

Additionally, we evaluate against 4D human reconstruction methods: AniGS~\cite{qiu2024anigs} and LHM~\cite{qiu2025lhm}. For AniGS, we obtain results using the official \href{https://modelscope.cn/studios/Damo_XR_Lab/Motionshop2}{API}. For LHM, we use the official implementation provided by the authors.

\vspace{\mylen}
\subsection{Novel View Synthesis}
\vspace{\mylen}

\noindent \textbf{Quantitative Comparison.}
We select 50 3D scans from the four multi-view datasets and render them into 360-degree static 3D videos to create our test set for comparison.  
The frontal view is used as the reference image.
During inference, our method samples exclusively along the view dimension as described in Section~\ref{sec:section_3}. 
As shown in Table~\ref{tab:3d_scan}, quantitative comparisons demonstrate that our method surpasses others, highlighting its effectiveness in learning precise 3D viewpoint transformations.

\begin{figure*}[h]
    \centering
\includegraphics[width=0.9\linewidth]{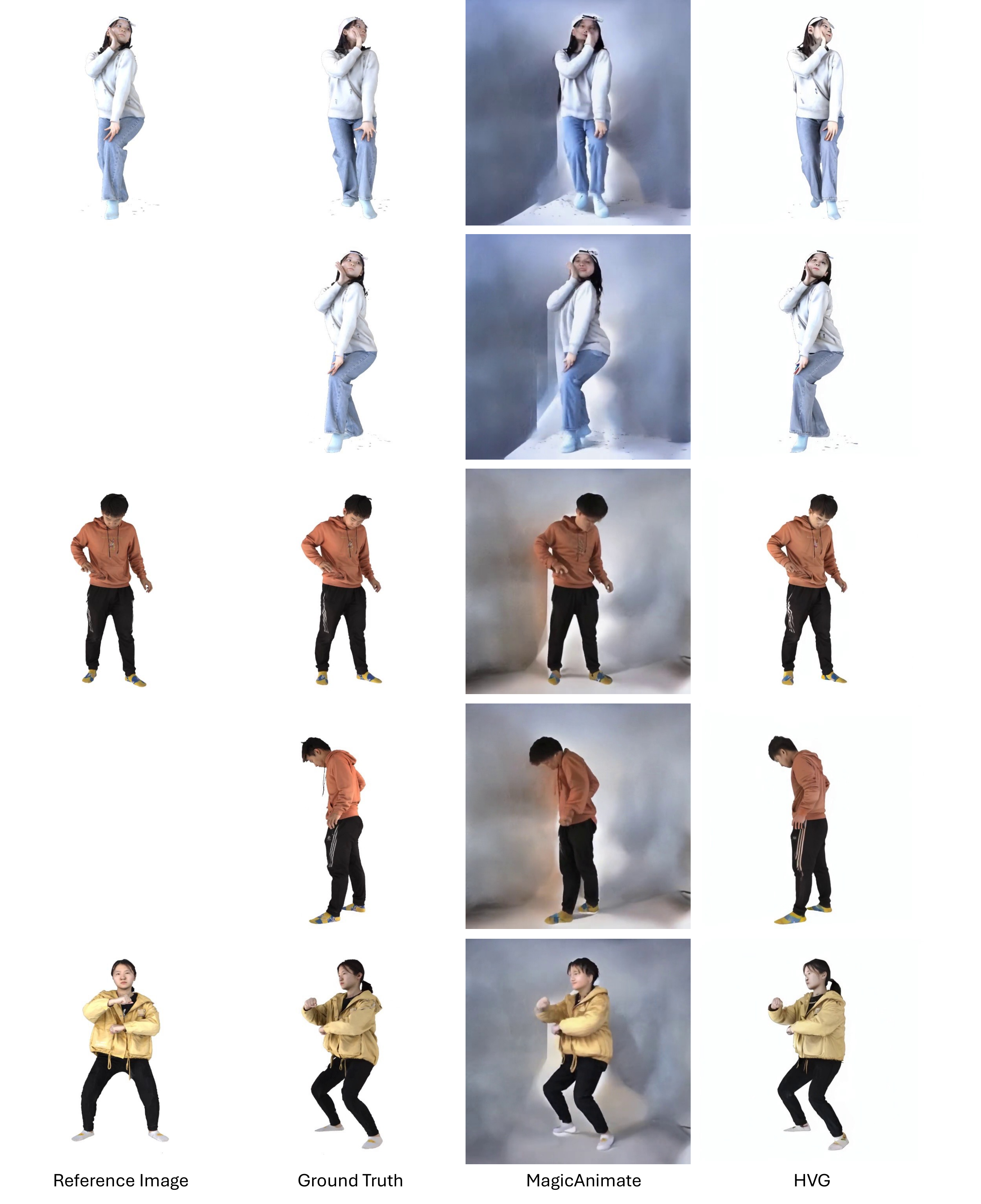}
    \vspace{-5pt}
        \caption{{\bf Novel-view results from single-view images.} \ourMethod avoids the head distortions produced by MagicAnimate and predicts textures closer to the ground truth.}
\label{fig:multi_view_magicanimate}
\end{figure*}

\begin{figure*}[tp]
    \centering
\includegraphics[width=\linewidth]{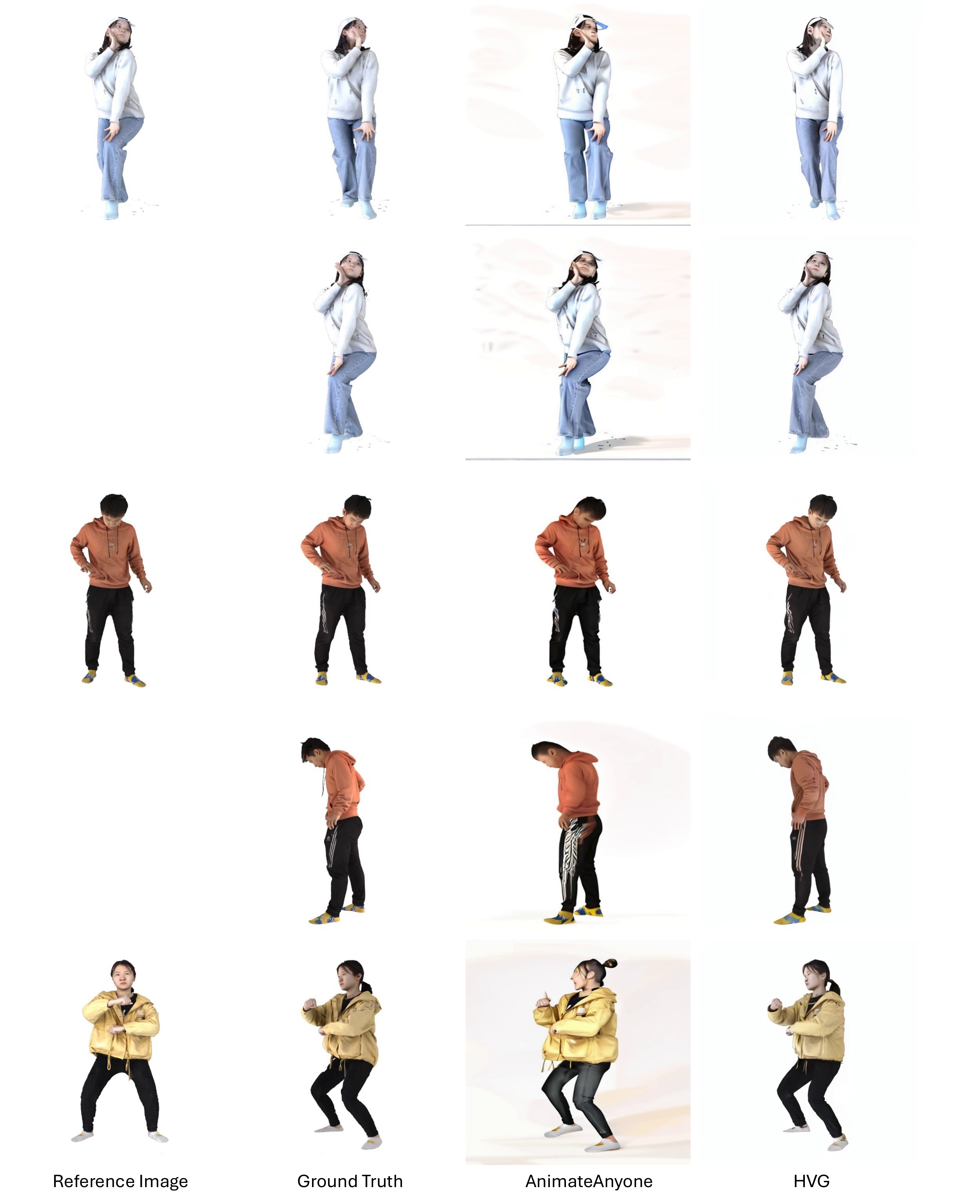}
    \vspace{-5pt}
        \caption{{\bf Novel-view results from single-view images.} \ourMethod avoids the head distortions produced by AnimateAnyone and predicts textures closer to the ground truth.}
\label{fig:multi_view_animateanyone}
\end{figure*}

\begin{figure*}[tp]
    \centering
\includegraphics[width=\linewidth]{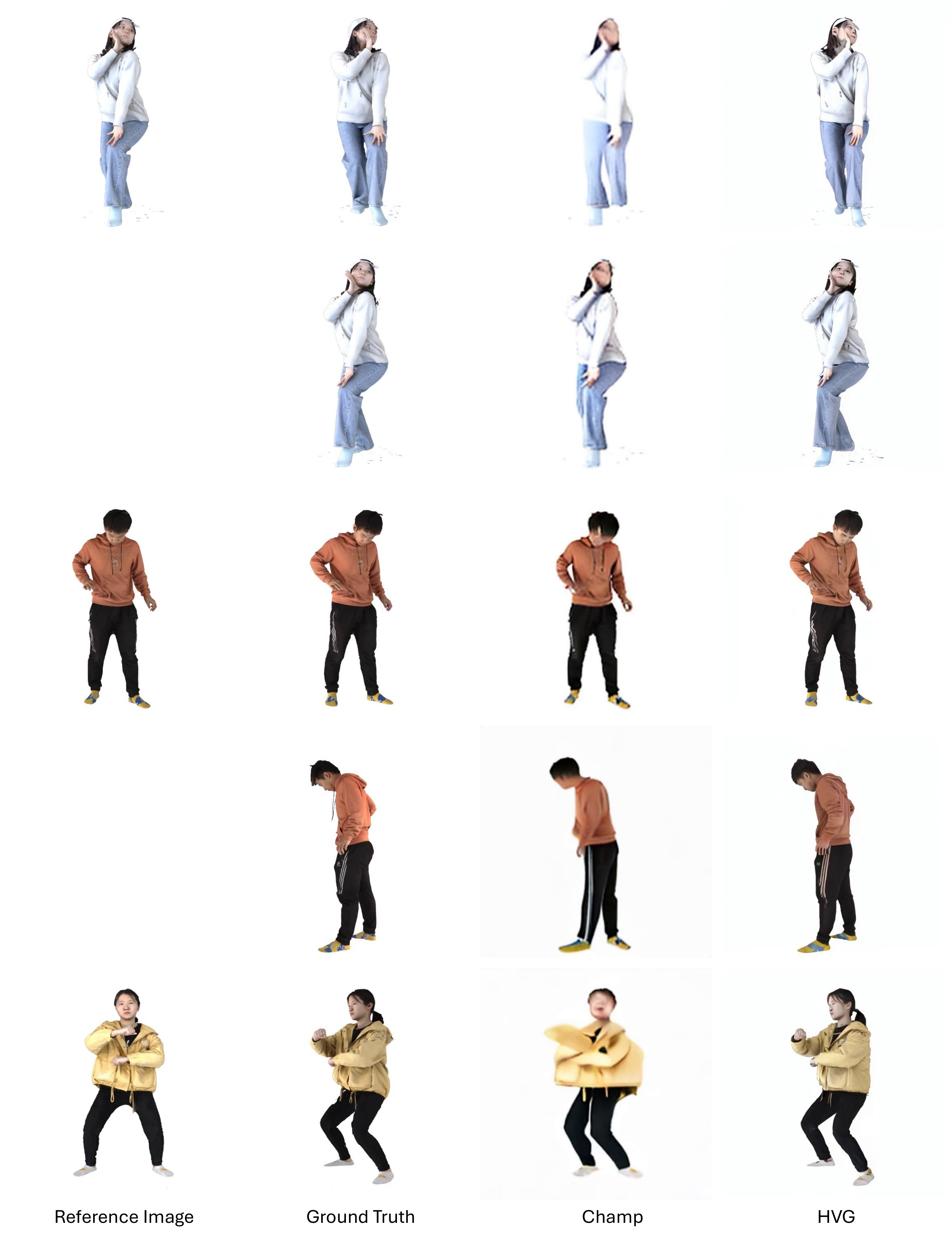}
    \vspace{-5pt}
        \caption{{\bf Novel-view results from single-view images.} Compared with Champ, \ourMethod eliminates head distortions and produces sharper, more detailed textures.}
\label{fig:multi_view_champ}
\end{figure*}

\begin{figure*}[tp]
    \centering
\includegraphics[width=\linewidth]{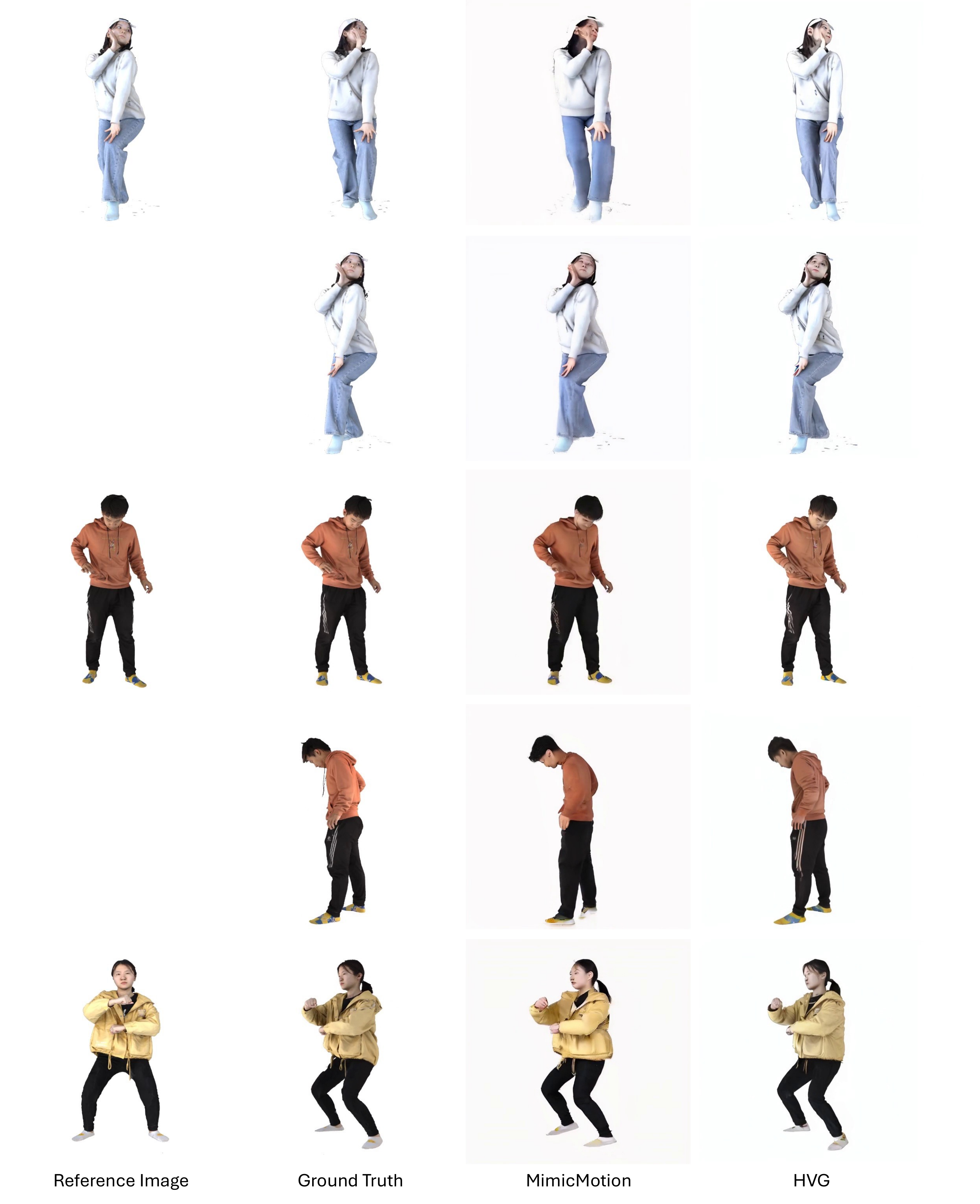}
    \vspace{-5pt}
        \caption{{\bf Novel-view results from single-view images.} Compared with MimicMotion, \ourMethod produces sharper, more detailed textures.}
\label{fig:multi_view_mimicmotion}
\end{figure*}

\begin{figure*}[tp]
    \centering
\includegraphics[width=\linewidth]{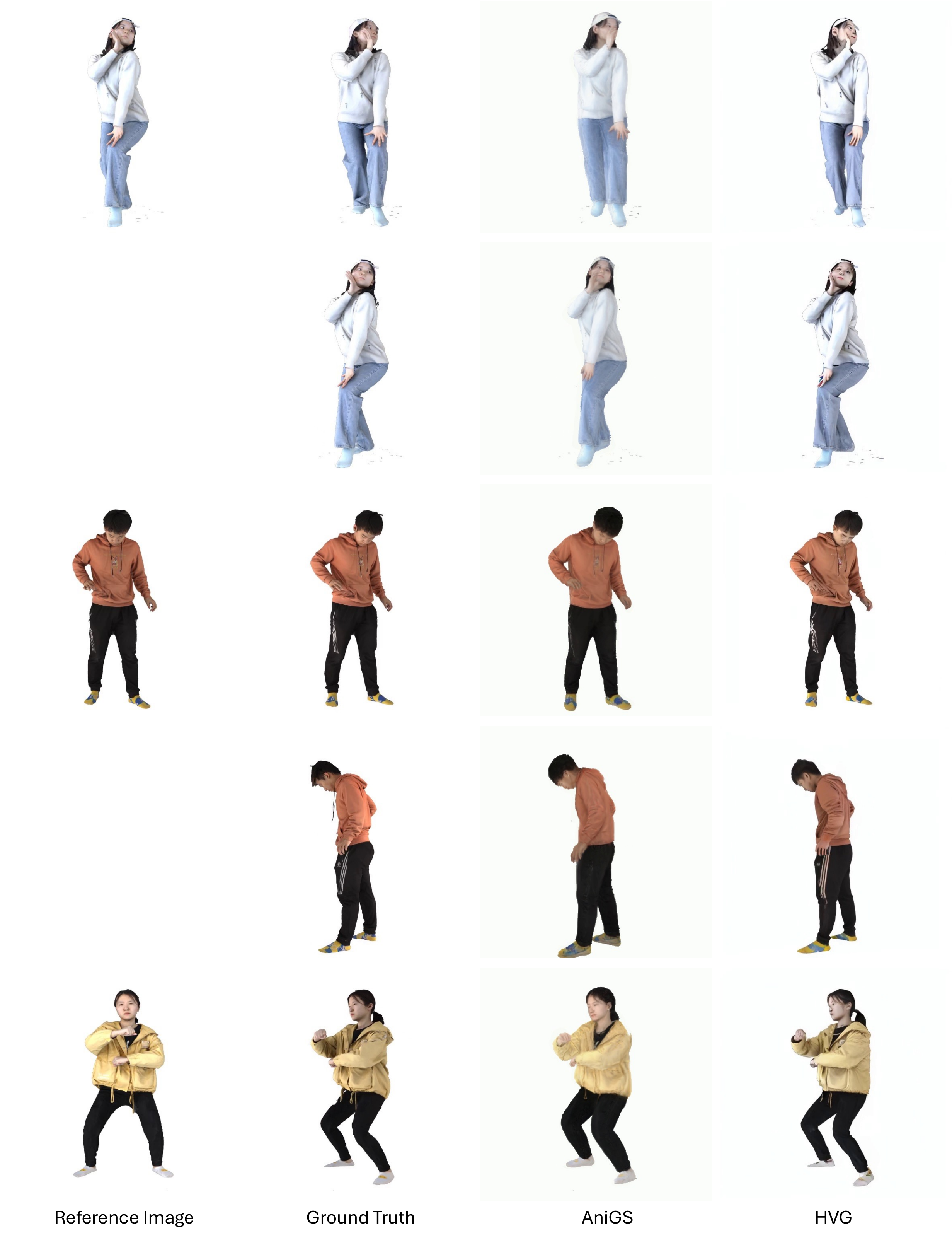}
    \vspace{-5pt}
        \caption{{\bf Novel-view results from single-view images.} AniGS struggles to generate motion-dependent textures, whereas \ourMethod effectively addresses this limitation and generates more clearer textures on the clothes.}
\label{fig:multi_view_anigs}
\end{figure*}

\begin{figure*}[tp]
    \centering
\includegraphics[width=\linewidth]{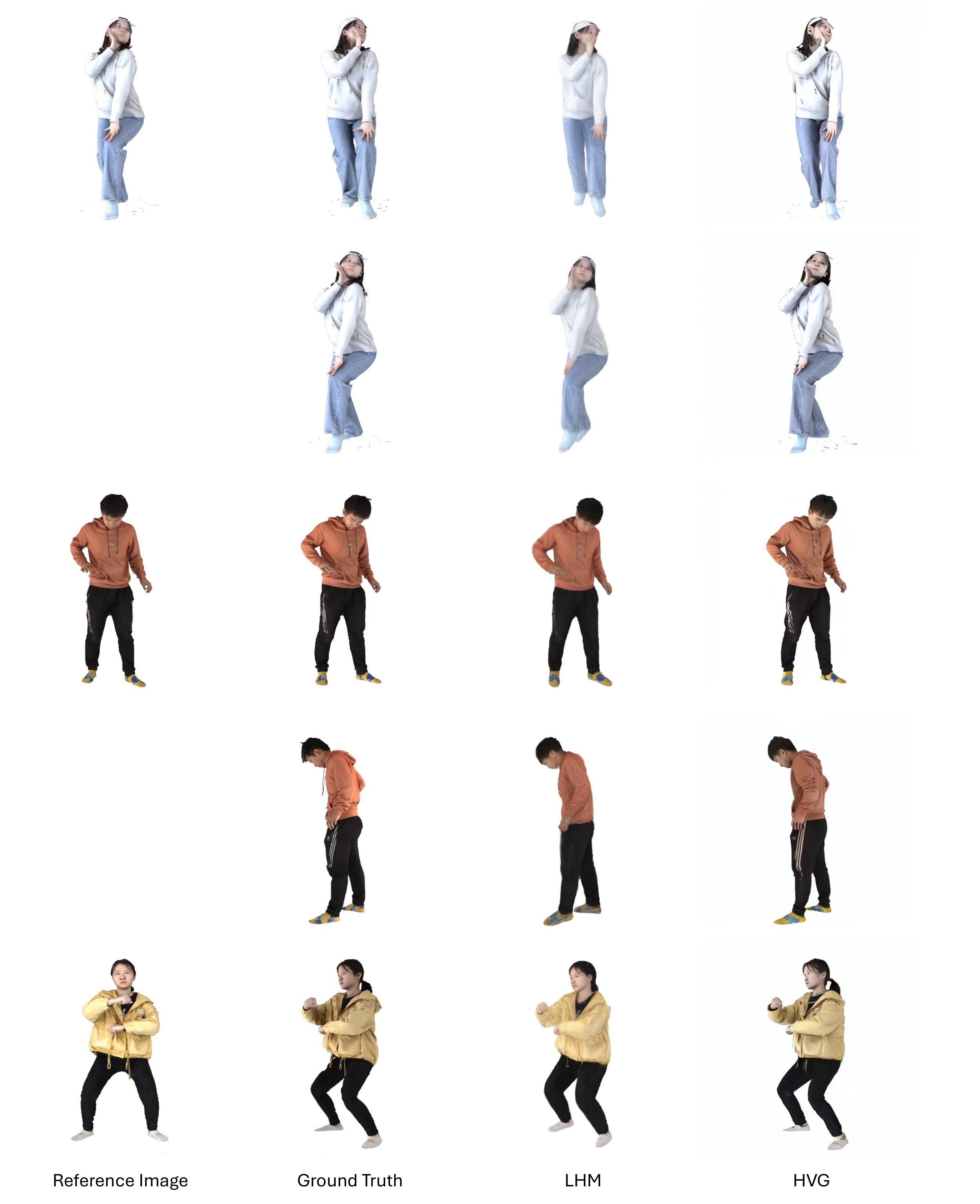}
    \vspace{-5pt}
        \caption{{\bf Novel-view results from single-view images.} LHM struggles to generate motion-dependent textures, whereas \ourMethod effectively addresses this limitation.}
\label{fig:multi_view_lhm}
\end{figure*}

\noindent \textbf{Qualitative Results.}
In Figure~\ref{fig:fig_visual_multi-view},~\ref{fig:multi_view_magicanimate},~\ref{fig:multi_view_animateanyone},~\ref{fig:multi_view_champ},~\ref{fig:multi_view_mimicmotion},~\ref{fig:multi_view_anigs},~\ref{fig:multi_view_lhm} and the supplementary video, we present a qualitative comparison between \ourMethod and several baselines. 
Notably, our method achieves superior multi-view coherence across diverse identities and poses.
Compared to existing approaches, it exhibits a stronger ability to maintain consistency across multiple viewpoints, effectively reducing artifacts and occlusion that commonly arise during human rotation.
As shown in the first and third rows, both AnimateAnyone~\cite{hu2024animate} and MimicMotion~\cite{zhang2024mimicmotion} produce distorted arms, whereas our approach avoids this issue. This advantage is attributed to the proposed bone map, which effectively addresses the occlusion problem.
Furthermore, compared to MimicMotion, which produces overly smooth textures on clothing, \ourMethod generates more detailed and realistic representations, demonstrating its effectiveness in preserving fine-grained appearance details.

\subsection{Novel View and Novel Pose Synthesis}
\vspace{\mylen}

\noindent \textbf{Quantitative Comparison.} 
For the evaluation of our multi-view video generation method, we randomly select 25 videos from the MVHumanNet dataset. 
We choose 8 different views, all captured at nearly the same height in world space, to cover a full 360-degree camera rotation around the subject. 
For each reference image, we generate videos with all 8 views, yielding a total of 200 videos. Here the views close to the frontal view as the primary reference image. During the inference phase, our method applies the proposed progressive spatio-temporal sampling strategy. This strategy leverages consistent spatial and temporal sampling across views, allowing our method to handle viewpoint transitions seamlessly.
As shown in Table~\ref{tab:3d_scan_quantitative}, quantitative comparisons indicate that our approach outperforms existing methods in managing both camera viewpoint and human motion variations concurrently. 
These results highlight our model’s robustness in dynamic 4D scenarios, demonstrating its capability to learn complex, realistic changes in both spatial orientation and temporal movement.

\begin{table}[tp]
    \centering
    \footnotesize
    \resizebox{0.5\textwidth}{!}{
    \begin{tabular}{lccc}
        \hline
        Method & PSNR~$\uparrow$ & LPIPS~$\downarrow$ & FVD~$\downarrow$ \\
        \hline
        \ourMethod (w/o normal map) &20.34 &0.068 &205.7\\
        \ourMethod (w/o depth map) &20.69  &0.063 &186.2
\\
        \ourMethod (w/ skeleton map) &18.52 &0.088 &277.5\\
        \ourMethod (w/ SMPL body normal map) &19.61 &0.073 &230.6 \\
        \ourMethod & 22.13 &0.057& 152.1 \\
        \hline
    \end{tabular}
    }
    \caption{{\bf Ablation study on bone maps.} Removing normal and depth maps leads to clear performance degradation, while combining these complementary cues yields the best results. Bone maps also outperform using only the skeleton map or the SMPL body normal map, underscoring their effectiveness in addressing occlusion and reducing shape-leakage artifacts. }
    \label{tab:video_comp}
\end{table}

\begin{figure*}[h]
\centering
\includegraphics[width=0.7\linewidth]{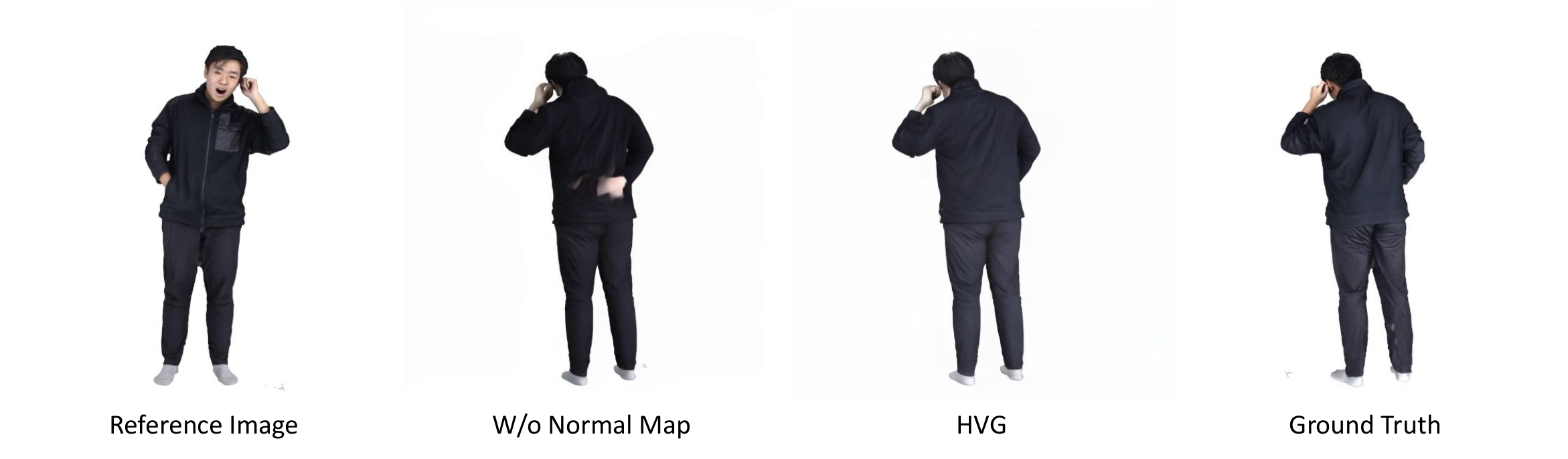}\\
\includegraphics[width=0.7\linewidth]{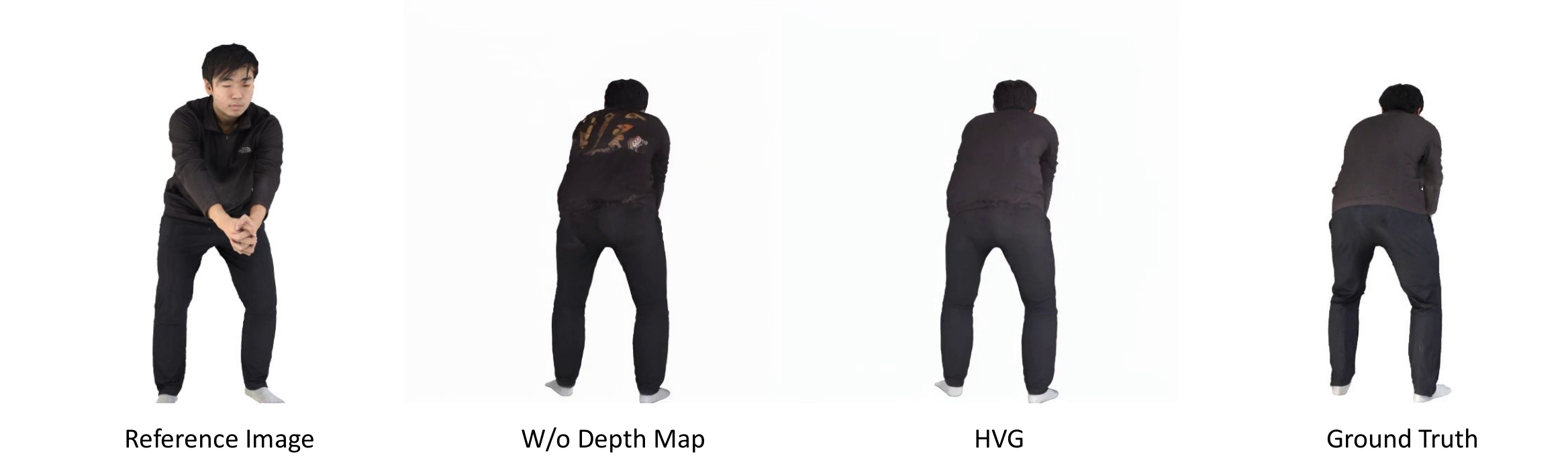}\\
\includegraphics[width=0.7\linewidth]{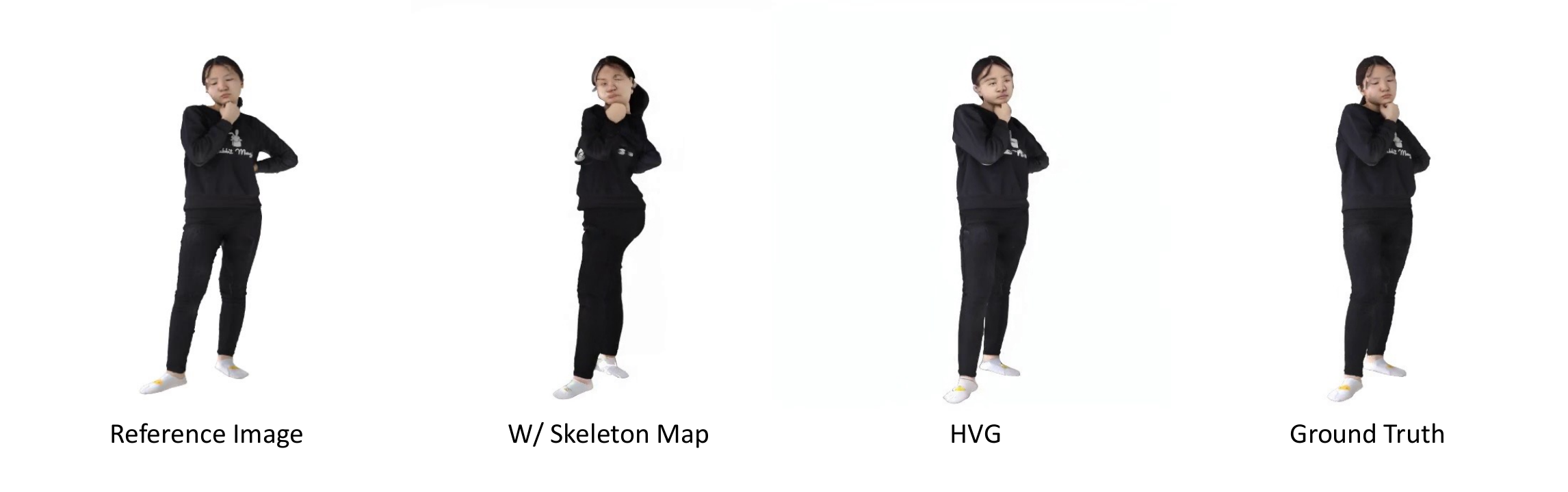}\\
\includegraphics[width=0.7\linewidth]{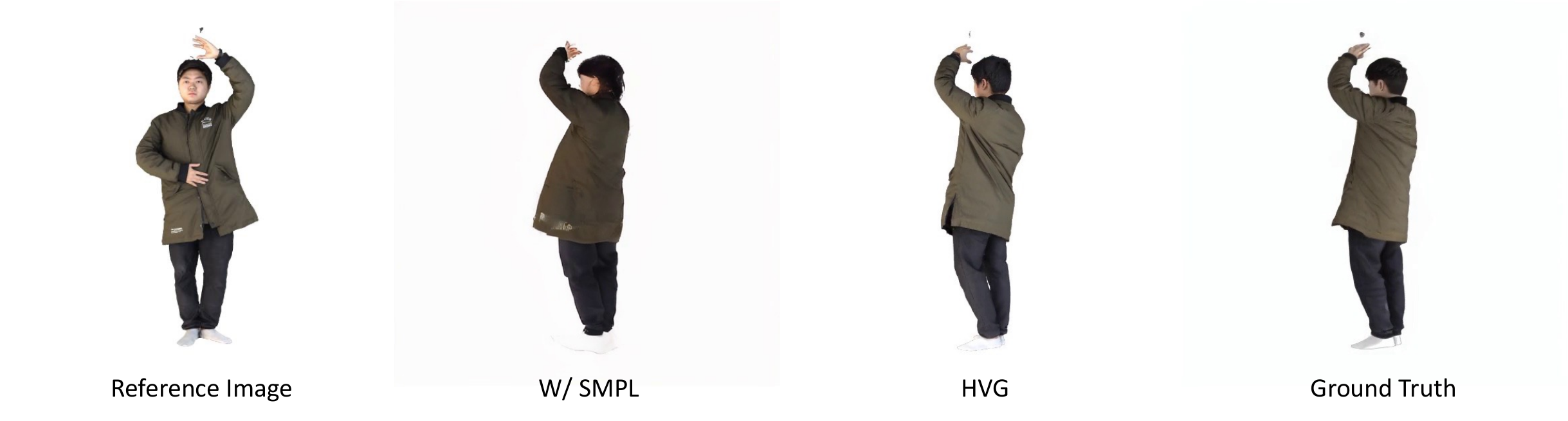}
\caption{{\bf Ablation study on bone maps.} Without either the normal or depth input, the generated human exhibits clear clothing artifacts, highlighting the importance of these geometric cues. Skeleton-only input causes limb misplacement, and SMPL-normal–only input leads to exaggerated clothing contours. Our full model \ourMethod resolves both artifacts by introducing the depth and normal bone maps.}
\label{fig:aba_normal_depth_skeleton_smpl}
\end{figure*}

\begin{figure*}[tp]
    \centering
\includegraphics[width=0.85\linewidth]{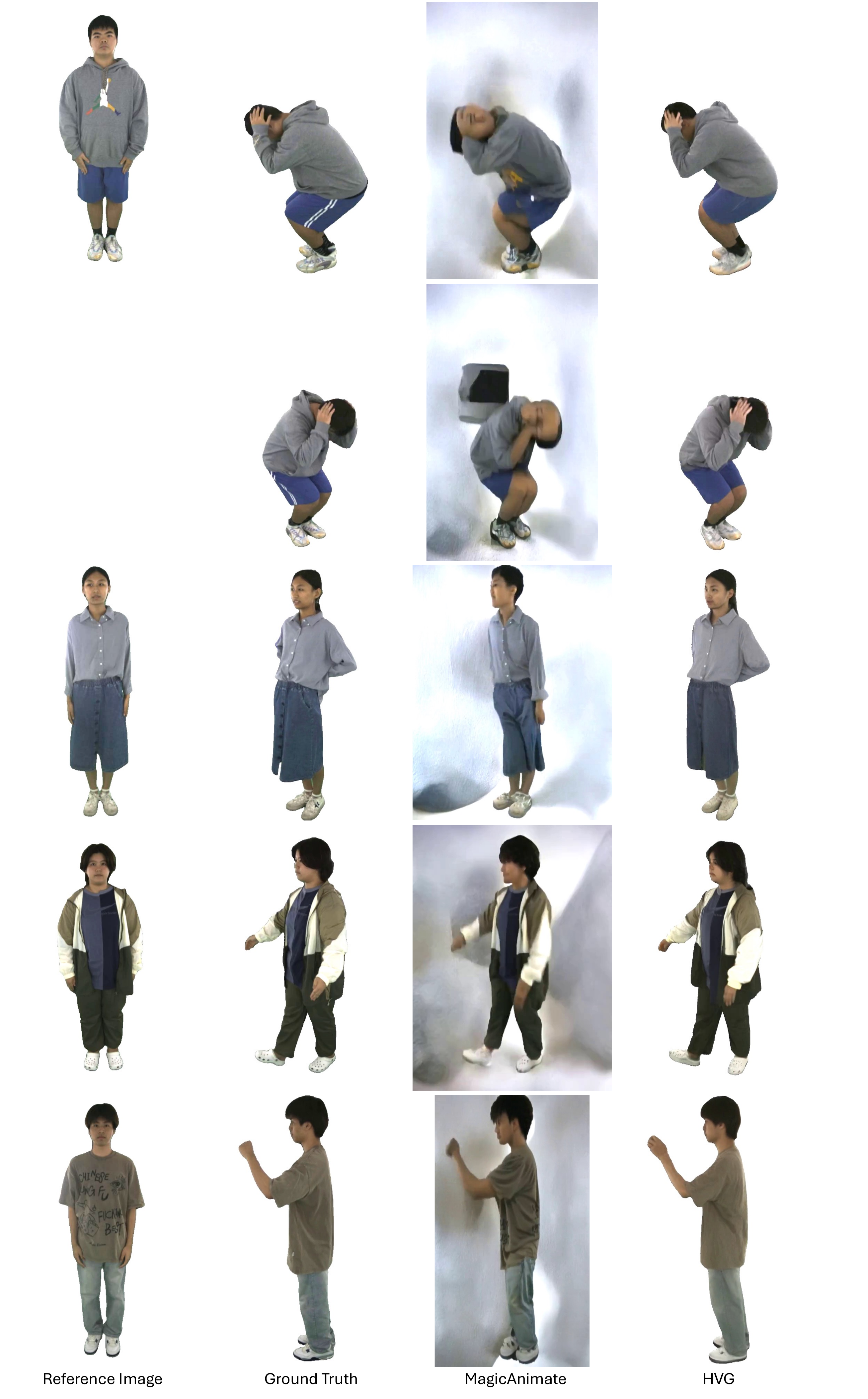}
    \vspace{-5pt}
        \caption{{\bf Novel-view novel-pose results from single-view images.} \ourMethod avoids the head distortions produced by MagicAnimate and predicts textures closer to the ground truth.}
\label{fig:multi_videos_magicanimate}
\end{figure*}

\begin{figure*}[tp]
    \centering
\includegraphics[width=0.85\linewidth]{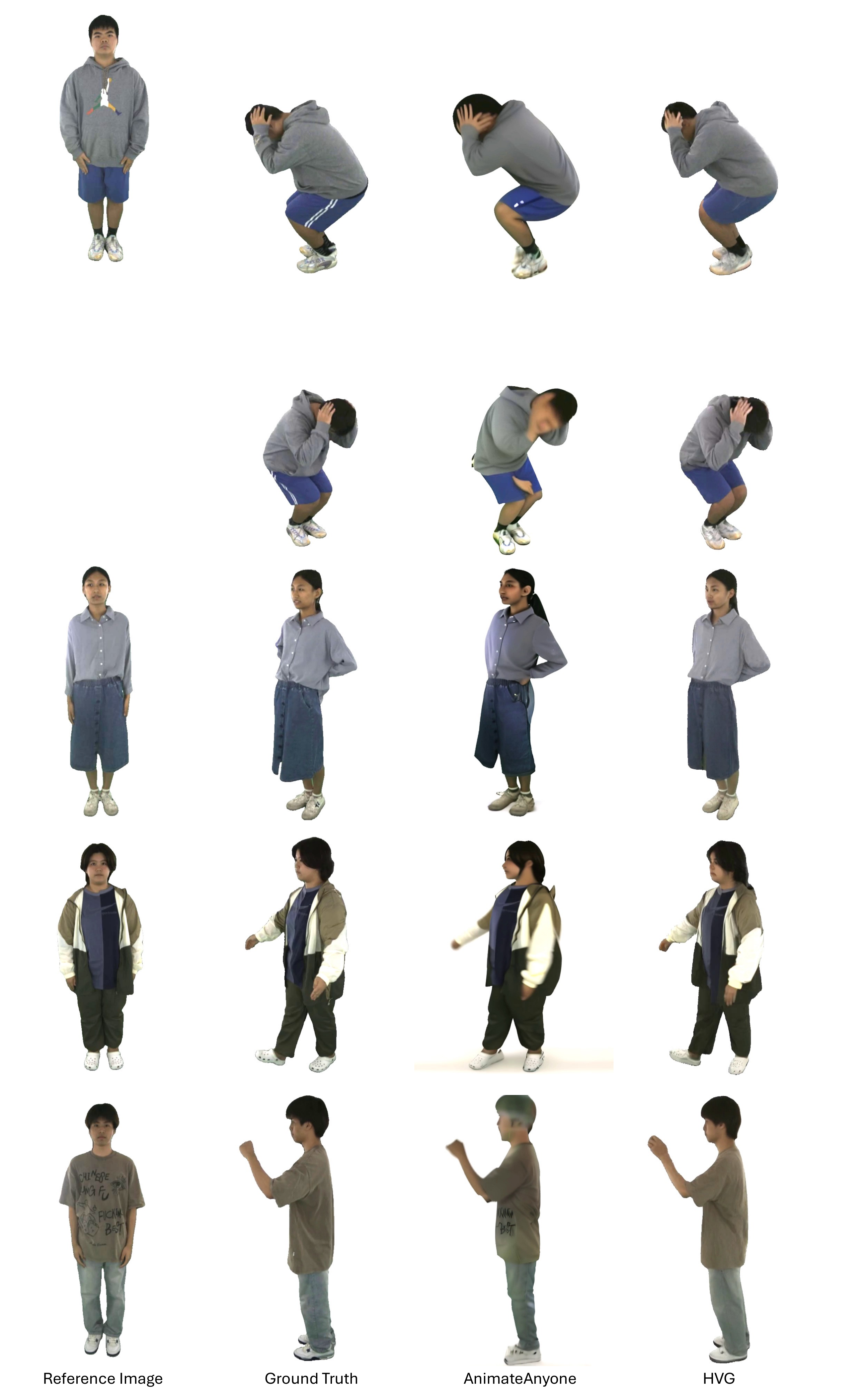}
    \vspace{-5pt}
        \caption{{\bf Novel-view novel-pose results from single-view images.} \ourMethod avoids the head distortions produced by AnimateAnyone and predicts textures closer to the ground truth.}
\label{fig:multi_videos_animateanyone}
\end{figure*}

\begin{figure*}[tp]
    \centering
\includegraphics[width=0.85\linewidth]{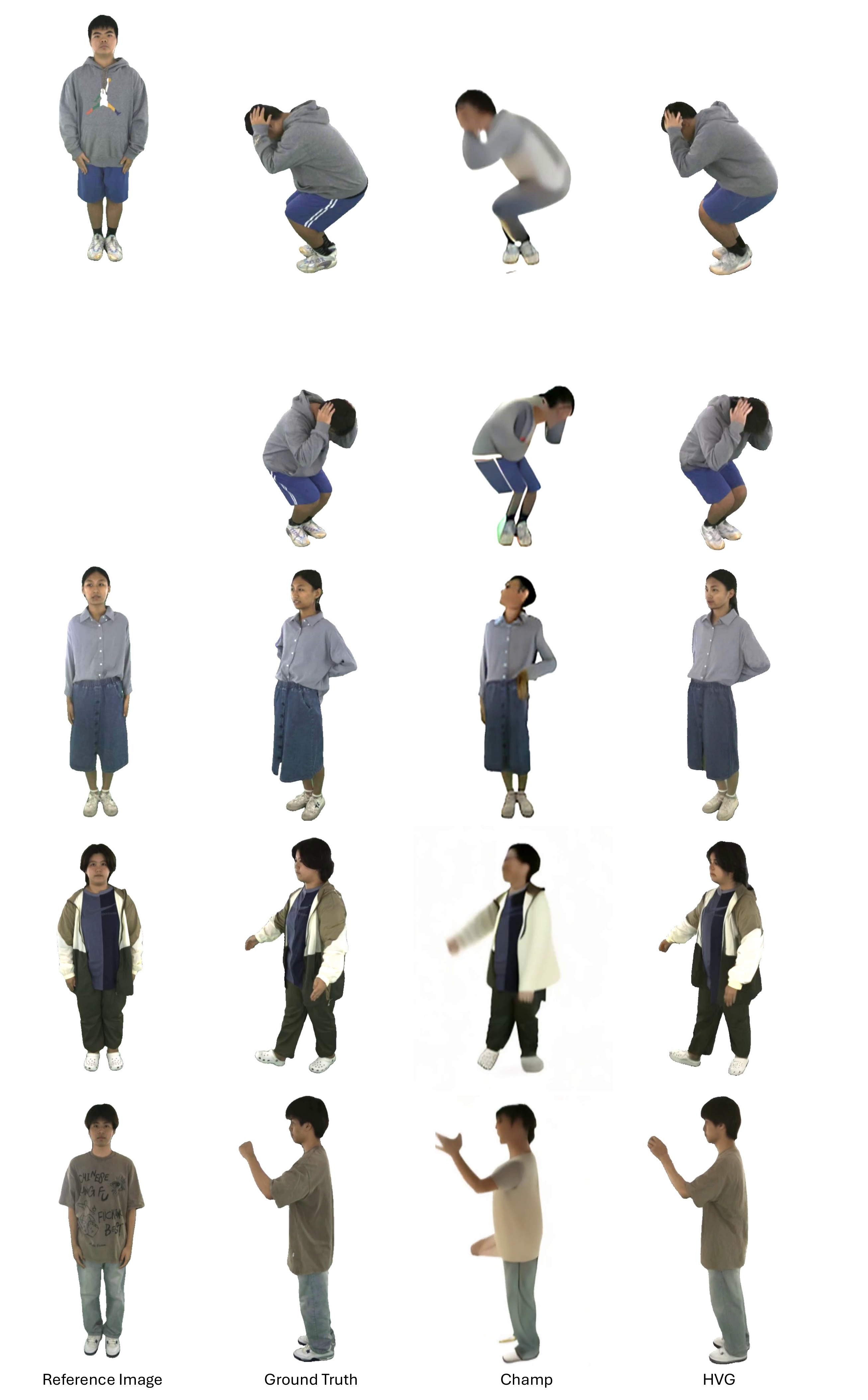}
    \vspace{-5pt}
        \caption{{\bf Novel-view novel-pose results from single-view images.} Compared with Champ, \ourMethod produces sharper, more detailed textures.}
\label{fig:multi_videos_champ}
\end{figure*}

\begin{figure*}[tp]
    \centering
\includegraphics[width=0.85\linewidth]{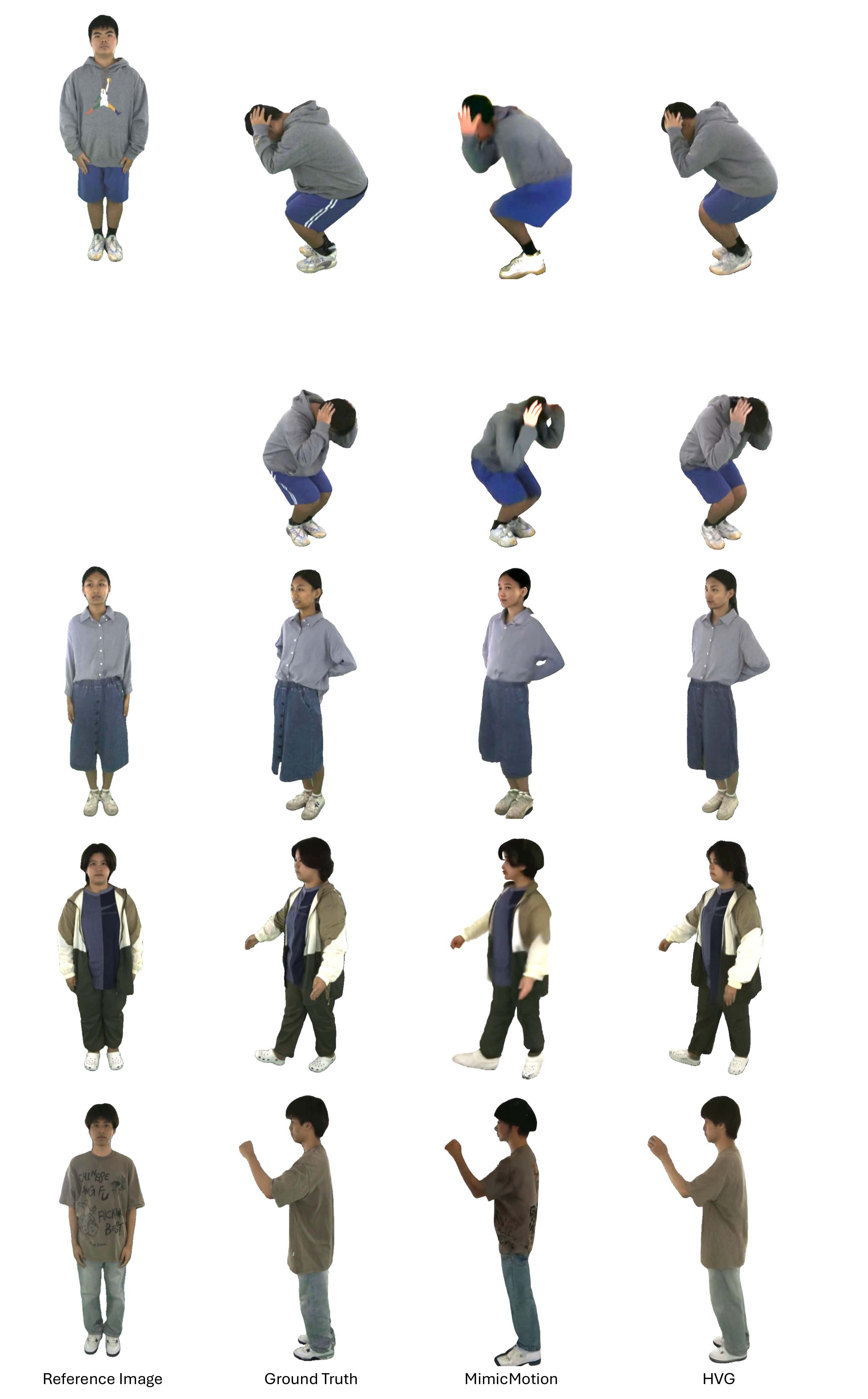}
    \vspace{-5pt}
        \caption{{\bf Novel-view novel-pose results from single-view images.} Compared with MimicMotion, \ourMethod produces sharper, more detailed textures.}
\label{fig:multi_videos_mimicmotion}
\end{figure*}

\begin{figure*}[tp]
    \centering
\includegraphics[width=0.85\linewidth]{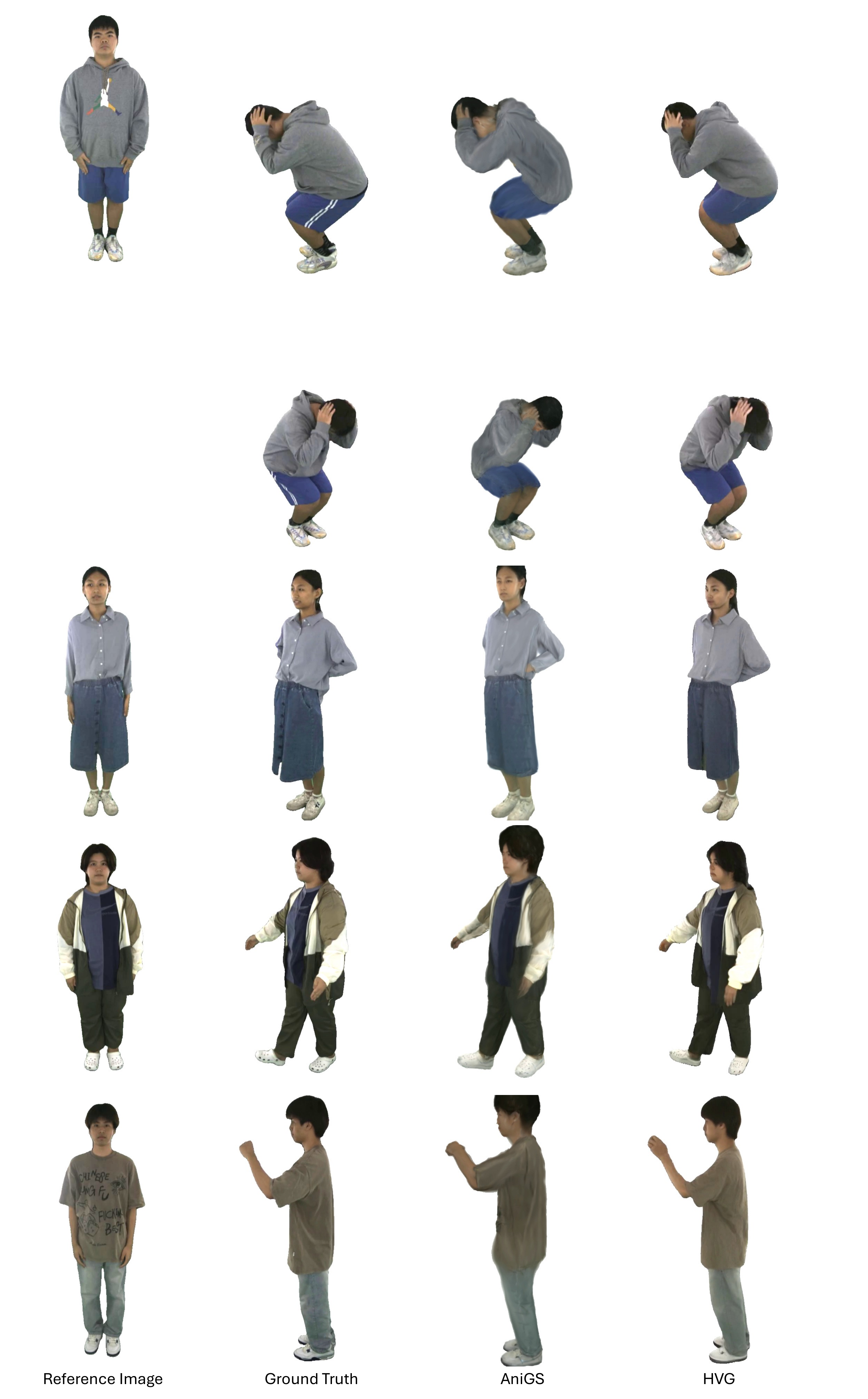}
    \vspace{-5pt}
        \caption{{\bf Novel-view novel-pose results from single-view images.} AniGS struggles to generate motion-dependent textures, whereas \ourMethod effectively addresses this limitation and generates more clearer textures on the clothes.}
\label{fig:multi_videos_anigs}
\end{figure*}

\begin{figure*}[tp]
    \centering
\includegraphics[width=0.85\linewidth]{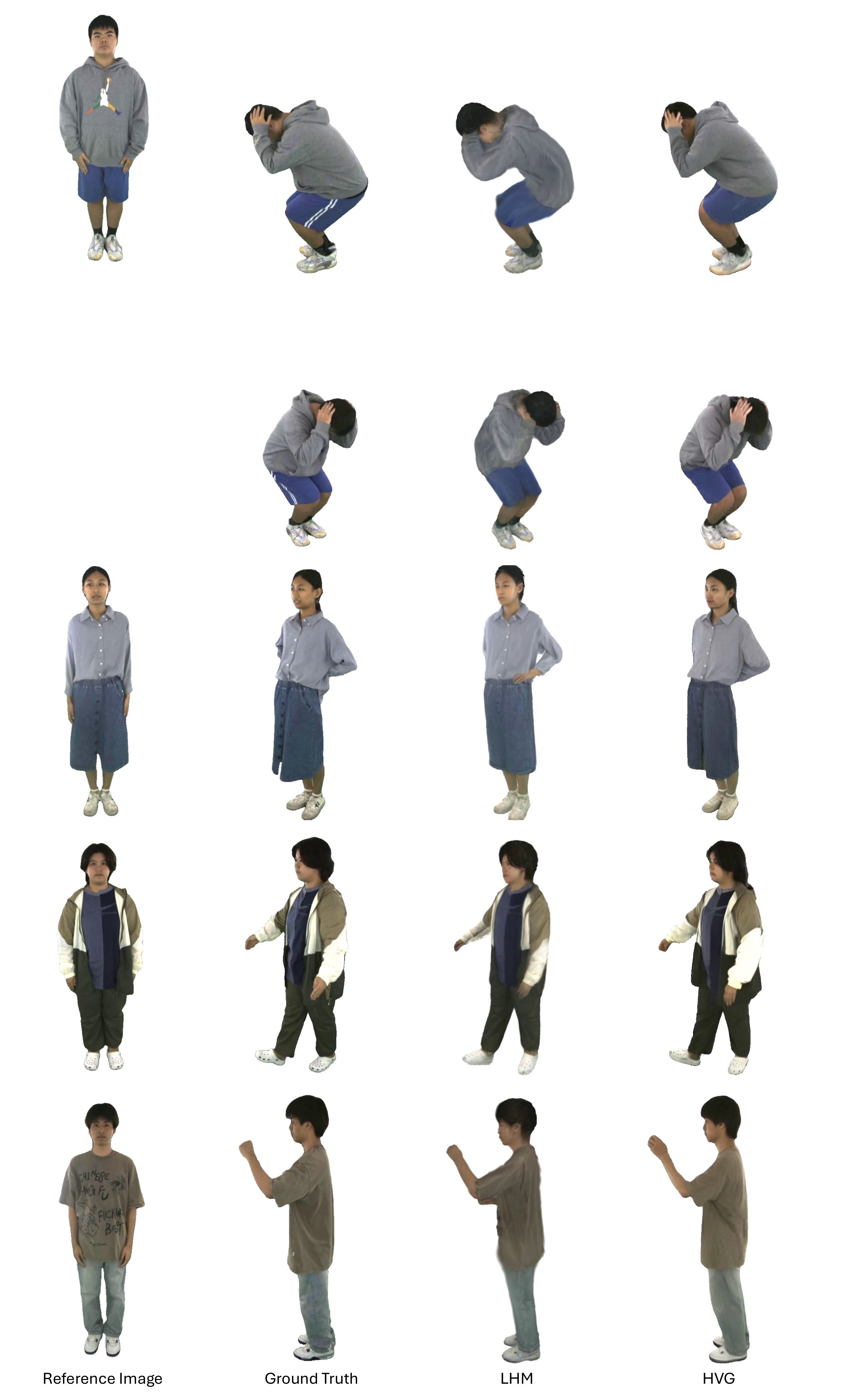}
    \vspace{-5pt}
        \caption{{\bf Novel-view novel-pose results from single-view images.} LHM struggles to generate motion-dependent textures, whereas \ourMethod effectively addresses this limitation and generates more clearer textures on the clothes.}
\label{fig:multi_videos_lhm}
\end{figure*}

\noindent \textbf{Qualitative Results.} 
As demonstrated in Figure~\ref{fig:fig_visual},~\ref{fig:multi_videos_magicanimate},~\ref{fig:multi_videos_animateanyone},~\ref{fig:multi_videos_champ},~\ref{fig:multi_videos_mimicmotion},~\ref{fig:multi_videos_anigs},~\ref{fig:multi_videos_lhm} and the supplementary video on the MVHumanNet dataset, our method, \ourMethod, showcases exceptional spatio-temporal consistency across multiple views while excelling in preserving high-fidelity details in challenging regions such as clothing textures, occluded regions during human turning. Specifically, in the third row of Figure~\ref{fig:fig_visual}, \ourMethod effectively captures the fine-grained texture changes of the coat during stretching. In contrast, baseline methods tend to overemphasize texture details, resulting in unrealistic and exaggerated clothing appearances. The second row highlights \ourMethod's ability to handle occlusion effectively, particularly in scenarios involving arm movements that partially obscure the torso; our method seamlessly reconstructs the occluded regions, including the arms and head, whereas AnimateAnyone and MagicAnimate struggle with them. 
These improvements are a direct result of \ourMethod's innovative components, that is the multi-input condition pose maps, which provide robust anatomical guidance to address occlusion and shape leakage problem.

\begin{figure*}[tp]
\vspace{-2mm}
    \centering
\includegraphics[width=\linewidth]{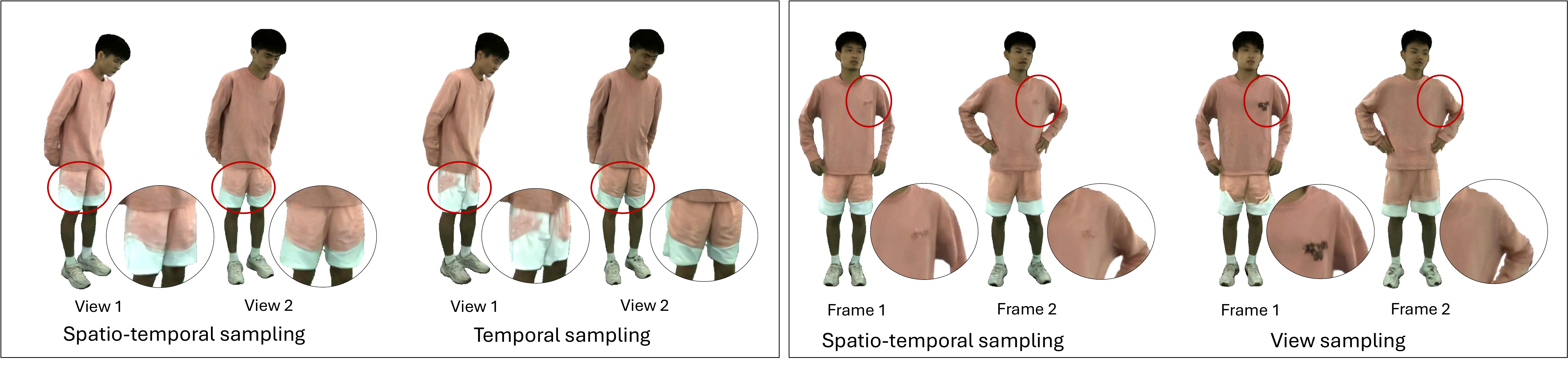}
    \caption{{\bf Effectiveness of our spatio-temporal sampling strategy.} We compare our results for two adjacent views with only temporal sampling in the left image. The right image shows results for two adjacent frames when only spatial window sampling is applied.}
\label{fig:sampling}
\end{figure*}

\begin{figure*}[h]
\centering
\includegraphics[width=0.8\linewidth]{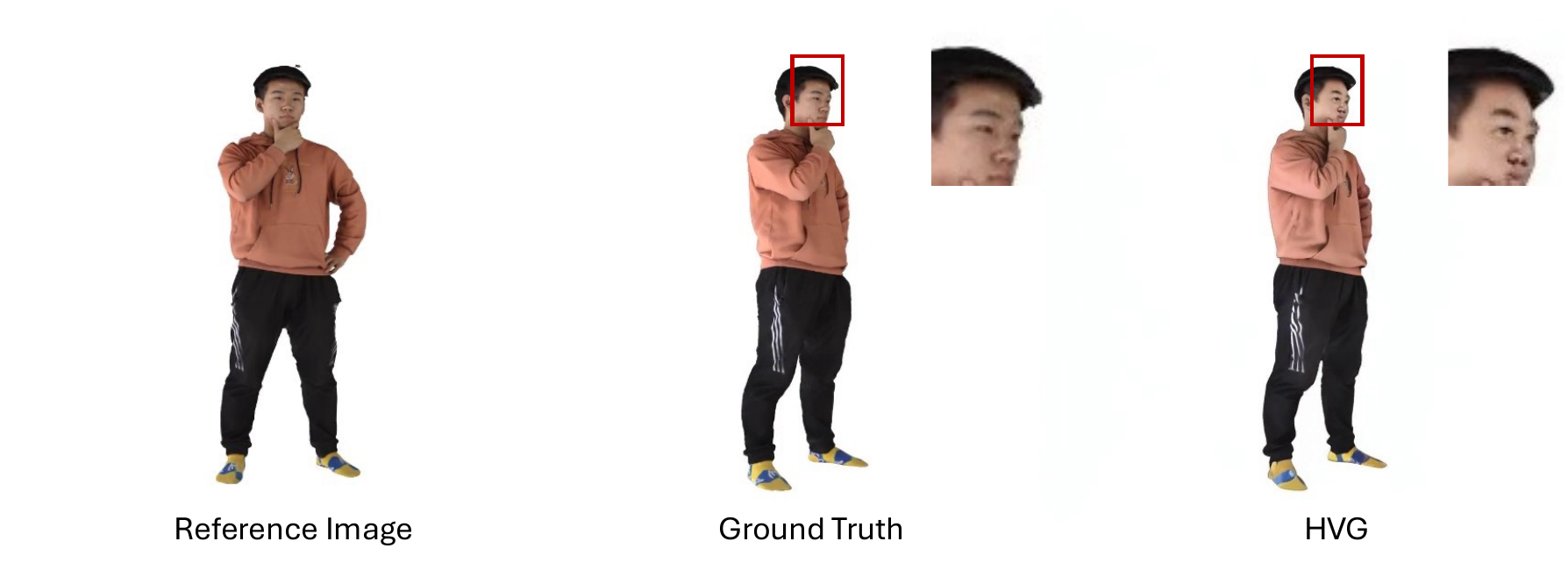}
        \caption{{\bf Failure case.} In generating full-body 4D human videos, \ourMethod may introduce artifacts in facial regions due to the inherent trade-off between capturing global structure and preserving local fidelity.}
\label{fig:failure_case}
\end{figure*}

\vspace{\mylen}
\subsection{Ablation Study}
\vspace{\mylen}

\textbf{Effectiveness of the Pose Modulator.}
Table~\ref{tab:video_comp} demonstrates that the full configuration of our proposed method~(“\ourMethod”) significantly outperforms other variants in terms of image quality, fidelity, video consistency, and realism. The ablation results for “w/o normal map” and “w/o depth map” indicate a decrease in performance when these components are removed. Notably, excluding normal and depth maps results in degraded performance, while combining the two complementary cues achieves the best overall results. 
This underscores the importance of their interaction in preserving shape alignment and motion guidance. Visual results can be found in the first and second rows of Figure~\ref{fig:aba_normal_depth_skeleton_smpl}.
Furthermore, as shown in the third and fourth rows of Table~\ref{tab:video_comp} and Figure~\ref{fig:aba_normal_depth_skeleton_smpl}, bone maps outperform the results only using skeleton map or SMPL body normal map, highlighting their advantage in handling occlusion issues and mitigating shape leakage problems.

\noindent \textbf{Effectiveness of Sampling Strategy.} 
To validate the effectiveness of our spatial-temporal sampling approach, we conduct a qualitative comparison, as shown in Figure~\ref{fig:sampling}. The left image presents results for two adjacent views when only temporal window sampling is applied, where noticeable inconsistencies appear (\eg the color of the shorts shifts from white to orange). The right image displays results for two adjacent frames when only view window sampling is used, revealing inconsistencies such as the appearance and disappearance of the T-shirt logo. 
In contrast, our sampling method mitigates these issues, achieving global coherence across the generated sequence. These results demonstrate that our approach effectively addresses the challenge of long-range dependencies in multi-view human video generation, enhancing both visual quality and coherence.

\section{Failure Case}
We focus on 4D human video generation of the full body, which may inadvertently introduce artifacts in finer facial regions due to the trade-off between global coherence and local detail. As illustrated in Figure~\ref{fig:failure_case}, our method occasionally produces distortions in areas such as the nose and lips, where precise geometry and texture are critical for realism. This issue arises because the current framework prioritizes consistent motion and structure across the entire body, potentially underrepresenting high-frequency facial details. A potential solution is to crop the head region and process it separately using a dedicated neural network specialized for facial generation. This modular approach allows finer control and resolution in the facial area, which can then be seamlessly fused with the body output to achieve higher overall visual fidelity.

\vspace{\mylen}
\section{Conclusion}
\label{sec:conclusion}
\vspace{\mylen}

In this paper, we introduced Human Video Generation in 4D~(\ourMethod), a novel latent video diffusion model designed to generate high-quality, spatiotemporally coherent human videos from a single image with 3D pose and pose control. 
By leveraging Articulated Pose Modulation, View and Temporal Alignment, and Progressive Spatio-Temporal Sampling, \ourMethod captures detailed, view-consistent clothing dynamics while ensuring smooth multi-view transitions. 
Our extensive experiments demonstrate that \ourMethod surpasses existing methods in pose-guided image-to-video synthesis.

\section*{Data Availability}
The datasets used in this study are publicly available and are accessible from their respective project websites and repositories cited in the references.

\end{sloppypar}

\bibliographystyle{unsrt}

\clearpage
\onecolumn

\section*{Appendix}
In this appendix, we present additional qualitative results to further illustrate the effectiveness of our approach.

\end{document}